\documentclass{article}

\PassOptionsToPackage{numbers,square,sort&compress}{natbib}
\usepackage[eandd,final]{neurips_2026}
\setcitestyle{numbers,square}

\usepackage[utf8]{inputenc}
\usepackage[T1]{fontenc}

\usepackage{float}
\usepackage{placeins}
\usepackage{url}
\usepackage{hyperref}
\hypersetup{hidelinks}
\usepackage{booktabs}
\usepackage{amsfonts}
\usepackage{amsmath}
\usepackage{amssymb}
\usepackage{nicefrac}
\usepackage{microtype}
\usepackage[table]{xcolor}
\usepackage{rotating}
\usepackage{pdflscape}
\usepackage{scalefnt}
\usepackage{graphicx}
\usepackage{multirow}
\usepackage{array}
\usepackage{tabularx}
\usepackage{makecell}
\usepackage{enumitem}
\usepackage{needspace}
\usepackage{pifont}
\usepackage{wrapfig}
\usepackage[most]{tcolorbox}
\usepackage{titlesec}
\usepackage[font=small,labelfont=bf]{caption}

\newtcolorbox{findingbox}{
  colback=black!2,
  colframe=black!65,
  boxrule=0.6pt,
  arc=2pt,
  left=5pt, right=5pt, top=2pt, bottom=2pt,
  fontupper=\small,
  before={\par\addvspace{3pt}\Needspace{10\baselineskip}\noindent},
  after={\par\addvspace{2pt}},
  enhanced jigsaw,
  breakable
}

\newcommand{\micon}[1]{%
  \raisebox{-0.12\height}{\includegraphics[height=0.85em]{figures/model_icons/#1.png}}\,}

\titlespacing*{\section}{0pt}{1.4ex plus 0.3ex minus 0.2ex}{0.7ex}
\titlespacing*{\subsection}{0pt}{1.0ex plus 0.2ex minus 0.1ex}{0.4ex}
\titlespacing*{\paragraph}{0pt}{0.4ex plus 0.2ex minus 0.1ex}{0.5em}
\setlist[itemize]{topsep=0.3ex, itemsep=0.2ex, parsep=0pt}
\setlist[enumerate]{topsep=0.3ex, itemsep=0.2ex, parsep=0pt}
\newcommand{\avrb}{\textsc{StemBind}}
\newcommand{\sac}{\textsc{Acc}}
\newcommand{\stc}{\textsc{StemCoherent}}
\newcommand{\vg}{\textsc{VG}}
\newcommand{\ssa}{\textsc{SSA}}
\newcommand{\tg}{\textsc{ThinkGain}}

\newcommand{\tgx}{\textsc{ThinkGain@X}}

\title{\textsc{StemBind}: When MLLMs Get Lost Between Rules and Instances in Abstract Visual Reasoning}

\author{%
  \textbf{Xixiang He}$^{1}$\thanks{Equal contribution.} \quad
  \textbf{Baiqi Wu}$^{2}$\footnotemark[1] \quad
  \textbf{Xingming Li}$^{1}$ \quad
  \textbf{Ao Cheng}$^{1}$\\
  \textbf{Qiyao Sun}$^{1}$ \quad
  \textbf{Xuanyu Ji}$^{1}$ \quad
  \textbf{Qingyong Hu}$^{3}$\thanks{Corresponding author.}\\[0.35em]
  {\normalfont\small $^{1}$National University of Defense Technology}\\
  {\normalfont\small $^{2}$Zhejiang University \quad
  $^{3}$Intelligent Game and Decision Lab}\\[0.35em]
  {\normalfont\footnotesize \texttt{\{hexixiang,lixingming,chengao18,sunqiyao18,jixuanyu18\}@nudt.edu.cn}}\\
  {\normalfont\footnotesize \texttt{wubaiqi@zju.edu.cn}, \texttt{huqingyong15@outlook.com}}\\
}

\begin{document}

\maketitle
\setcounter{footnote}{0}

\vspace{-1.6em}
{\centering\small
\href{https://hexixiang.github.io/StemBind/}{\micon{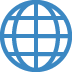}Project Page}\hspace{1.8em}%
\href{https://github.com/hexixiang/StemBind}{\micon{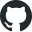}Code}\hspace{1.8em}%
\href{https://huggingface.co/datasets/user48271/Stembind}{\micon{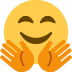}Dataset}\par}
\vspace{0.3em}

\begin{abstract}
Multimodal large language models (MLLMs) often \emph{know the rule but pick the wrong answer}: on abstract visual reasoning (AVR) tasks, a model can correctly describe what it sees and correctly name the underlying pattern, yet still fail to choose the matching candidate. Existing AVR benchmarks cannot detect this gap because they collapse perception, rule induction, and answer selection into a single right-or-wrong signal. We introduce \textbf{\avrb{}}, a shared-stem diagnostic benchmark that probes the same visual stem with three aligned questions: \textbf{P}erception (what is in the image), \textbf{R}ule (what pattern governs it), and \textbf{F}ull (which option completes it), so that a final-answer error can be attributed to a specific sub-step on the same evidence. \avrb{} contains \textbf{2{,}298} curated knowledge-light stems across nine auditable visual operations, totaling \textbf{19{,}533} P/R/F tasks, with each full item annotated by Sternberg's four reasoning stages: \textbf{S1 Encode}, \textbf{S2 Infer}, \textbf{S3 Map}, and \textbf{S4 Apply}. Evaluating \textbf{24 frontier MLLM configurations} (proprietary and open-source) yields four findings. \textbf{(i)~The R--F chasm.} Rule accuracy exceeds full-item accuracy on \textbf{22 of 24} models, so most failures happen \emph{after} the rule has been identified. \textbf{(ii)~A persistent binding gap.} Even when P and R are both correct on the same stem, models still answer F incorrectly \textbf{51.2\%} of the time. \textbf{(iii)~The bottleneck is S3.} Process diagnostics and Stage-wise Stimulus Augmentation (\ssa{}) localize the dominant failure to \emph{rule-to-instance mapping}, the step that binds an inferred rule to the right candidate. \textbf{(iv)~Scaling and thinking do not help.} Neither scaling up model size nor enabling explicit thinking mode reliably closes the gap; in paired comparisons, thinking mode lifts perception but \emph{lowers} both rule and full-item accuracy. \avrb{} reframes AVR evaluation from final-answer ranking to locating \emph{where} abstract visual reasoning breaks down, and identifies rule-to-instance binding as a concrete next target for vision-grounded reasoning.
\end{abstract}

  \section{Introduction}
  \label{sec:intro}
  
  \begin{figure}[!t]
    \centering
    \includegraphics[width=\textwidth]{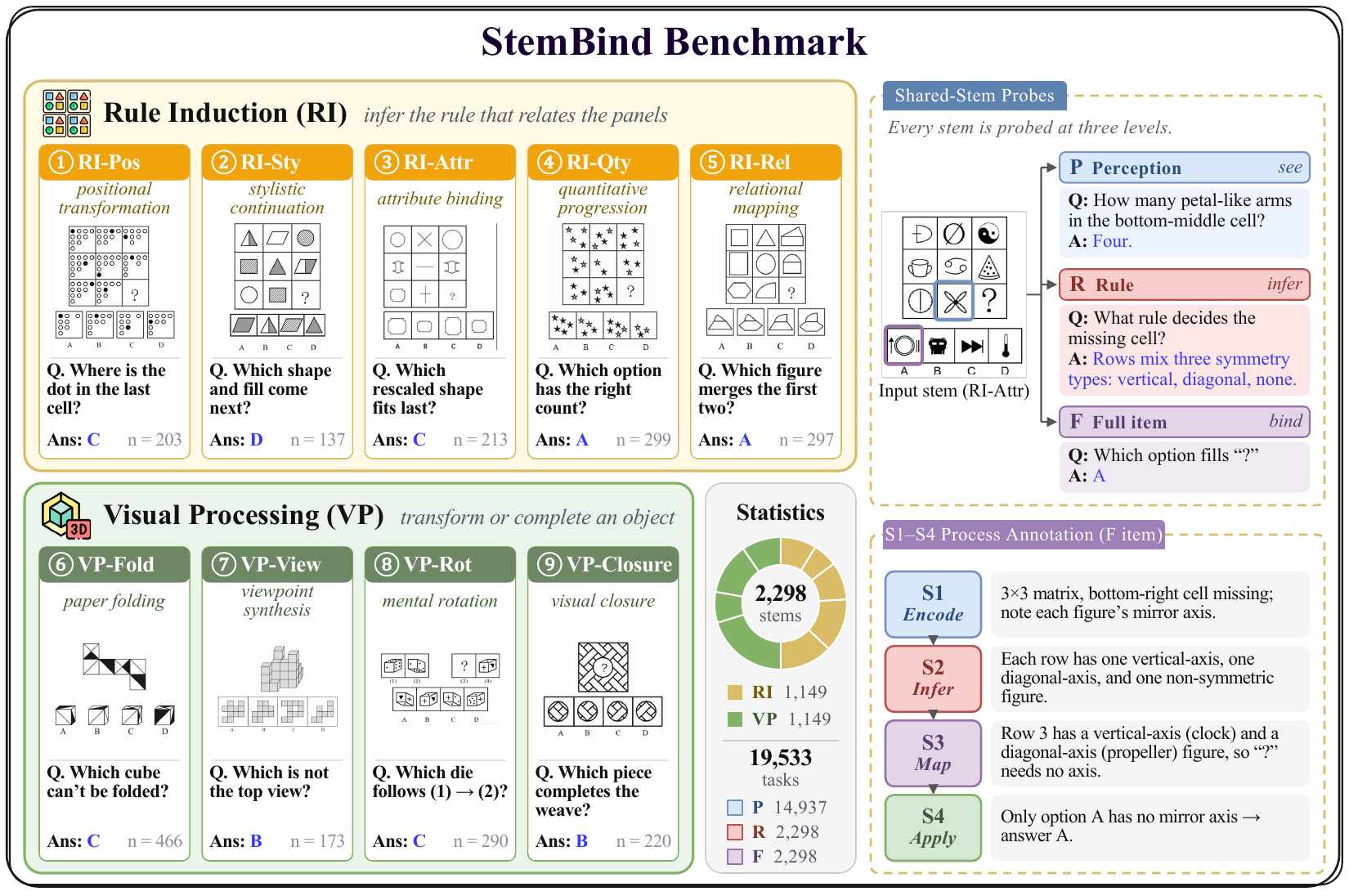}
    \caption{\avrb{} overview: 9 RI/VP operations, shared-stem P/R/F probes, and S1--S4 process stages. P, R, and F probes share the same visual stem; S1--S4 annotates the F item's solution path.}
    \label{fig:overview}
    \label{fig:taxonomy}
  \end{figure}

  Imagine a multimodal model that looks at a $3{\times}3$ visual matrix, correctly describes the shapes in every cell, correctly identifies the underlying rule as ``positional cycling,'' and confidently picks a candidate whose object sits in the wrong cell. From the perspective of a final-answer benchmark, this looks like a single failure indistinguishable from one where the model never saw the image at all. From a cognitive perspective, it is something quite different: the model has done the seeing and the thinking, yet \emph{cannot bind the rule it knows to the answer it must choose}. We argue that this kind of failure is not a corner case but a dominant mode of how today's multimodal large language models (MLLMs) fail at abstract visual reasoning, and no existing AVR benchmark can isolate it on the same visual stem.
  
  Abstract visual reasoning (AVR), exemplified by Raven-style procedural matrix datasets and visual concept benchmarks~\citep{raven,pgm,bongard_logo,conceptarc,arc}, is designed to isolate visual reasoning from math, physics, and world knowledge, the very confounds that broader benchmarks like MathVista and MMMU mix in~\citep{mathvista,mmmu,scienceqa,mathverse,olympiadbench}. Yet most AVR benchmarks~\citep{marvel,puzzlevqa,visualpuzzles,visulogic,visuriddles,vriq,iqbench,mmiq,morse500} score only the final answer, collapsing three very different failure modes into a single bit: did the model \emph{see} the elements? did it \emph{infer} the rule? did it \emph{bind} the rule to the right candidate? A few diagnostic benchmarks~\citep{visresbench,multistar} add perception or rule-style subquestions, but they pose those subquestions on \emph{different} stems from the full item, which means one can never ask the within-stem question that matters most: \emph{when a model has provably seen the elements and provably identified the rule, can it still get the answer right?}

  We build \avrb{} to ask that question (Figure~\ref{fig:overview}).~Each stem is probed three ways on the \emph{same} visual evidence: \textbf{P}erception (what is in the image), \textbf{R}ule (what pattern governs it), and \textbf{F}ull (which candidate completes it). To localize where a wrong answer comes from, every full item is further annotated with four reasoning stages drawn from Sternberg's componential theory of induction~\citep{sternberg1977}: \textbf{S1 Encode}, \textbf{S2 Infer}, \textbf{S3 Map}, and \textbf{S4 Apply}. These four stages map onto perception, rule induction, rule-to-instance alignment, and answer selection. \avrb{} also supports Stage-wise Stimulus Augmentation (\ssa{}), a controlled intervention that injects verified stage content cumulatively to test \emph{which} stage's information actually repairs a failed item. The result is \textbf{2{,}298} knowledge-light stems across nine auditable visual operations, expanded into \textbf{19{,}533} shared-stem P/R/F tasks with full S1--S4 annotations. \avrb{} is designed not as a larger leaderboard but as a diagnostic instrument.

  Evaluating \textbf{24 frontier MLLM configurations} (proprietary and open-source) under direct prompting shows a consistent pattern. \textbf{(i) The R--F chasm.} Rule accuracy exceeds full-item accuracy on \textbf{22 of 24} models, often by 20--34 points. \textbf{(ii) A persistent binding gap.} Even on the strict subset of stems where P and R are \emph{both} correct, models still answer F incorrectly \textbf{51.2\%} of the time. \textbf{(iii) The bottleneck is S3.} Both stage-wise judging and \ssa{} localize the dominant failure to rule-to-instance mapping, the moment when an inferred rule must be translated into a choice. \textbf{(iv) Scaling and thinking do not help.} Scaling models up and turning on explicit thinking mode do not close the gap; in paired comparisons, thinking mode even \emph{lowers} both rule and full-item accuracy. The picture that emerges is one in which today's MLLMs, on AVR tasks, often \emph{know the rule but lose it at the instance level}. Our main contributions are summarized as follows:

  \textbf{(i) A shared-stem diagnostic benchmark.} We release \avrb{}, the first AVR benchmark that probes perception, rule, and full-item solving on the \emph{same} visual stem, with \textbf{2{,}298} curated stems, nine knowledge-light visual operations, and \textbf{19{,}533} P/R/F tasks annotated with S1--S4 reasoning stages.

  \textbf{(ii) A process-aware error-attribution protocol.} We combine shared-stem probes, S1--S4 process targets, \ssa{} interventions, and paired direct/thinking controls~\citep{visualthoughts,visualsketchpad} into a single pipeline that attributes final-answer errors to specific sub-steps.

  \textbf{(iii) A binding-centered empirical diagnosis.} Across 24 direct-mode configurations, we show that the dominant residual failure of frontier MLLMs is rule-to-instance binding, supported by the R--F chasm, a strict conditional Binding Gap, and S3 localization that holds across model families.

  \textbf{(iv) Evidence against common fixes.} Under the same protocol, we show that neither model scaling nor explicit thinking mode reliably repairs the binding gap, identifying rule-to-instance binding as a concrete next target for vision-grounded reasoning.

  \section{Related Work}
  \label{sec:related}
  
  \textbf{MLLM visual reasoning benchmarks.} Broad or domain-specific benchmarks~\citep{mathvista,mmmu,mathverse,olympiadbench} mix visual reasoning with language priors, world knowledge, or domain conventions, while knowledge-light AVR benchmarks~\citep{marvel,puzzlevqa,visualpuzzles,visulogic,mmiq} better isolate abstract patterns. Diagnostic benchmarks~\citep{visresbench,multistar} add P/R-style probes or staged metrics, but none tie perception, rule, and full questions to the same visual stem. A fuller list of adjacent MLLM and AVR benchmarks is in Appendix~\ref{app:related_extended}.

  \textbf{Cognitive and psychometric benchmarks.} Raven-style and procedural matrix datasets~\citep{raven,iraven,pgm,bongard_logo,conceptarc,arc} study controlled rule induction; VISFACTOR~\citep{visfactor}, M3GIA~\citep{m3gia}, and SpatialViz-Bench~\citep{spatialviz} organize visual reasoning around FRCT, CHC, or spatial-visualization taxonomies. We use CHC Gf/Gv~\citep{mcgrew2009,schneider2012} only as coverage guidance, not as psychometric narrow-ability claims.

  \textbf{Process-aware and mode-aware evaluation.} Recent benchmarks add diagnostic signals through perception/reasoning labels, hierarchy-aware scoring, process judging, or thinking-mode analysis~\citep{lens,mmmath,adaptmmbench,visualprm}. Adjacent multi-image and sequence benchmarks, and vision-aware reasoning methods on natural images, videos, or math, are discussed in Appendix~\ref{app:related_extended}. \avrb{} brings shared-stem P/R/F probes, S1--S4 process targets, \ssa{}, and paired direct/thinking controls into controlled AVR.

\definecolor{clOurs}{RGB}{253,239,218}
\definecolor{clYes}{RGB}{51,160,44}
\definecolor{clNo}{RGB}{227,26,28}
\definecolor{clPart}{RGB}{255,153,0}
\newcommand{\yes}{\textcolor{clYes}{\ding{51}}}
\newcommand{\no}{\textcolor{clNo}{\ding{55}}}
\newcommand{\partly}{\textcolor{clPart}{\ding{108}}}

\newcolumntype{C}[1]{>{\centering\arraybackslash}p{#1}}
\newcolumntype{L}[1]{>{\raggedright\arraybackslash}p{#1}}
\newcolumntype{Y}{>{\raggedright\arraybackslash}X}
\newcommand{\bhdr}[1]{\textbf{\scriptsize #1}}

\begin{table}[H]
  \centering
  \caption{Benchmark comparison over diagnostic axes. \yes{} present, \partly{} partial, \no{} absent.}
  \label{tab:benchmark_comparison}
  \scriptsize
  \setlength{\tabcolsep}{1.8pt}
  \renewcommand{\arraystretch}{0.98}
  \begin{tabularx}{\textwidth}{@{}L{0.105\textwidth}L{0.135\textwidth}L{0.135\textwidth}
    *{8}{>{\centering\arraybackslash}X}@{}}
    \toprule
    \textbf{Family} & \textbf{Benchmark} & \textbf{Setting}
      & \bhdr{K-light}
      & \bhdr{Tax.}
      & \bhdr{Attr.}
      & \bhdr{Rule}
      & \bhdr{Stem}
      & \bhdr{Proc.}
      & \bhdr{Cue}
      & \bhdr{Mode} \\
    \midrule
    \multirow{7}{=}{AVR / visual reasoning}
      & MARVEL~\citep{marvel} & AVR puzzles
        & \yes & \yes & \yes & \no & \no & \no & \no & \partly \\
      & PuzzleVQA~\citep{puzzlevqa} & abstract patterns
        & \yes & \yes & \partly & \yes & \no & \partly & \yes & \no \\
      & VisualPuzzles~\citep{visualpuzzles} & visual puzzles
        & \yes & \yes & \no & \no & \no & \no & \no & \yes \\
      & VisuLogic~\citep{visulogic} & visual logic
        & \yes & \yes & \partly & \no & \no & \no & \no & \no \\
      & VisuRiddles~\citep{visuriddles} & abstract riddles
        & \yes & \yes & \yes & \partly & \no & \partly & \partly & \no \\
      & VRIQ~\citep{vriq} & IQ-style AVR
        & \yes & \yes & \yes & \yes & \partly & \no & \no & \partly \\
      & VisRes Bench~\citep{visresbench} & controlled visual
        & \yes & \yes & \partly & \partly & \no & \no & \no & \no \\
    \midrule
    \multirow{3}{=}{Cognitive / psychometric}
      & VISFACTOR~\citep{visfactor} & FRCT cognition
        & \partly & \yes & \partly & \no & \no & \no & \no & \partly \\
      & M3GIA~\citep{m3gia} & CHC multimodal
        & \no & \yes & \no & \no & \no & \no & \no & \no \\
      & SpatialViz-Bench~\citep{spatialviz} & spatial visualization
        & \yes & \yes & \partly & \partly & \no & \no & \no & \partly \\
    \midrule
    \multirow{2}{=}{Process / multi-input}
      & LENS~\citep{lens} & real-image tiers
        & \no & \yes & \yes & \partly & \partly & \partly & \no & \no \\
      & MMRB~\citep{mmrb} & multi-image reasoning
        & \no & \yes & \no & \yes & \no & \partly & \no & \partly \\
    \midrule
    \rowcolor{clOurs}
    \textbf{Ours}
      & \textbf{\avrb{}} & shared-stem AVR
        & \yes & \yes & \yes & \yes & \yes & \yes & \yes & \yes \\
    \bottomrule
  \end{tabularx}
\end{table}

\section{\avrb}
\label{sec:bench}

\subsection{Taxonomy and shared-stem diagnostic design}
\label{sec:design}

\paragraph{Design considerations.} A diagnostic AVR benchmark must answer not only ``which model is better'' but ``which sub-step of visual reasoning fails''. \textbf{(G1)~Knowledge-light stems}: items avoid math, physics, and world-knowledge content so errors more likely reflect visual reasoning. \textbf{(G2)~Theory-motivated, operational taxonomy}: CHC Gf/Gv motivates the RI/VP split, but each label is defined by reproducible visual operations rather than psychometric narrow-ability claims. \textbf{(G3)~Shared-stem diagnostic probes}: each stem is reused across perception, rule, and full-item probes so a final-answer error is comparable with perception and rule outcomes on the same visual evidence. \textbf{(G4)~Auditable contamination defense}: items are provenance-screened and served through an evaluation server with answer-mapping randomization. Together, (G1)--(G4) make \avrb\ a diagnostic instrument, not a larger AVR item pool.

\paragraph{9 knowledge-light operations.} Figure~\ref{fig:taxonomy} shows one representative stem per operation. The Rule-Induction (RI) family covers \textsc{RI-Pos} (positional transformation), \textsc{RI-Sty} (stylistic continuation), \textsc{RI-Attr} (attribute binding), \textsc{RI-Qty} (quantitative progression), and \textsc{RI-Rel} (relational mapping); the Visual-Processing (VP) family covers \textsc{VP-Fold} (paper-folding), \textsc{VP-View} (viewpoint synthesis), \textsc{VP-Rot} (mental rotation), and \textsc{VP-Closure} (visual closure). Each label is defined by the operation the solver must infer and a boundary rule separating it from neighboring types; legitimacy comes from operational definitions and annotation audits rather than psychometric norming. The operational-definition table and CHC reference map are in Appendix~\ref{app:chc_map}.

\begin{wrapfigure}{r}{0.48\textwidth}
  \vspace{-2.0ex}
  \centering
  \includegraphics[width=0.46\textwidth]{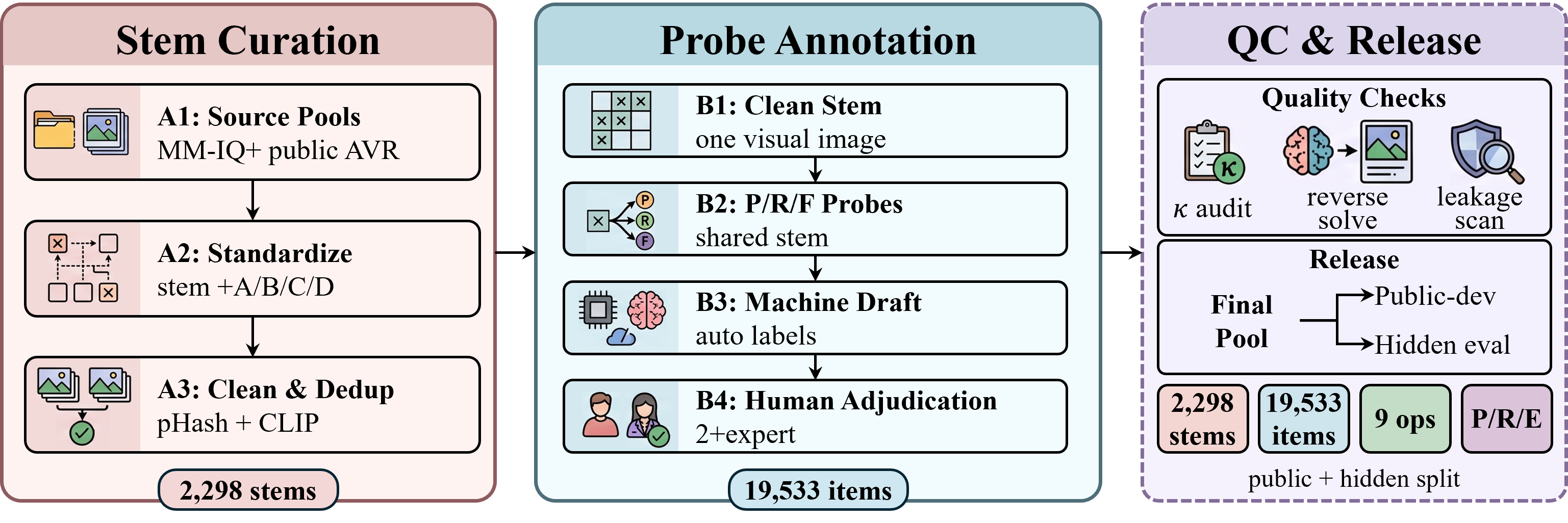}
  \caption{\avrb{} construction pipeline. Source pools are standardized and deduplicated into 2{,}298 stems, expanded into 19{,}533 shared-stem P/R/F tasks through machine drafting and human adjudication, and released after quality, leakage, and split checks.}
  \label{fig:pipeline}
  \vspace{-1.5ex}
\end{wrapfigure}

\paragraph{Shared-stem P/R/F probes.} Every stem instantiates three probe types sharing the same visual stem and annotation scaffold, but not a fixed number of instances. P (Perception) asks whether a model can describe the visual content; R (Rule), whether it selects the correct rule; F (Full), the original AVR item solved end-to-end. In the released split, each stem has one F item, one R probe, and on average $6.50$ P probes. Each P probe is derived from the gold S1 encoding of its stem. F-wrong while P and R are correct is consistent with a behavioral binding bottleneck; P-wrong with R correct suggests rule-guessing; both wrong indicates encoding collapse. The L2 evaluator and the \ssa\ intervention (\S\ref{sec:eval_protocol}) operationalize this cross-level logic.

\paragraph{S1--S4 process annotation.} Each F item carries four per-step targets following a Sternberg decomposition: S1 Encode (scene graph), S2 Infer (rule set), S3 Map (cross-panel alignment), and S4 Apply (target prediction). A perception-load tag (perception-heavy, rule-heavy, or mixed), distinct from probe level, feeds L3 attribution (\S\ref{sec:eval_protocol}). An expanded worked example with the full S1--S4 trace is in Appendix~\ref{app:worked_example}.

\subsection{Data curation and dataset analysis}
\label{sec:data}

\Needspace{18\baselineskip}
\begin{wraptable}{r}{0.45\textwidth}
  \vspace{-1.5ex}
  \centering
\caption{Key statistics of \avrb.}
\label{tab:dataset_overview}
\scriptsize
\setlength{\tabcolsep}{3.0pt}
\renewcommand{\arraystretch}{0.92}
\begin{tabular}{@{}p{3.50cm}>{\raggedleft\arraybackslash}p{2.45cm}@{}}
  \toprule
  \textbf{Statistics} & \textbf{Number} \\
  \midrule
  \rowcolor{black!5}
  \multicolumn{2}{@{}l@{}}{\textbf{Overall scale}} \\
  Stems & 2,298 \\
  Full evaluation tasks & 19,533 \\
  Operation types & 9 \\
  \midrule
  \rowcolor{black!5}
  \multicolumn{2}{@{}l@{}}{\textbf{Shared-stem probe coverage}} \\
  \quad Full items (F) & 2,298 (1.00 / stem) \\
  \quad Rule probes (R) & 2,298 (1.00 / stem) \\
  \quad Perception probes (P) & 14,937 (6.50 / stem) \\
  \midrule
  \rowcolor{black!5}
  \multicolumn{2}{@{}l@{}}{\textbf{Task hierarchy}} \\
  \quad Rule-Induction (RI) & 1{,}149 (50.0\%) \\
  \quad Visual-Processing (VP) & 1{,}149 (50.0\%) \\
  \midrule
  \rowcolor{black!5}
  \multicolumn{2}{@{}l@{}}{\textbf{Difficulty tags (Easy/Med./Hard)}} \\
  \quad Full items (F) & 4.0 / 77.0 / 19.0\% \\
  \quad Rule probes (R) & 8.4 / 89.7 / 1.8\% \\
  \quad Perception probes (P) & 48.6 / 33.8 / 17.6\% \\
  \midrule
  \rowcolor{black!5}
  \multicolumn{2}{@{}l@{}}{\textbf{Annotations}} \\
  S1--S4 process targets & 2,298 stems \\
  Difficulty tags & 19,533 tasks \\
  \bottomrule
\end{tabular}

  \vspace{-0.8ex}
\end{wraptable}

\paragraph{Curation, annotation, and contamination defense.} \avrb\ is a curated, contamination-audited AVR benchmark: released stems are retained only after provenance screening, near-duplicate filtering, and human review. Figure~\ref{fig:pipeline} summarizes the pipeline. Of the 2{,}298 stems, 1{,}011 (44.0\%) are newly constructed and 1{,}287 (56.0\%) are adapted from four license-screened public AVR datasets. Adapted stems keep only the abstract rule and answer logic; their visuals and options are rebuilt, and all probes and S1--S4 targets are newly annotated (Appendix~\ref{app:curation_pipeline}). pHash + CLIP deduplication with dual-VLM pre-screening removes near-duplicates; GPT-5 and Claude Opus propose options and S1--S4 targets only as a draft stage. Two trained annotators review every item, an expert adjudicator resolves disagreements, and a daily 10\% golden-set agreement check monitors drift. Two frozen VLMs re-solve each item as a leakage probe, and option-text shortcuts are filtered~\citep{agrawal2018,geirhos2020}. The result-bearing split is released directly; the hidden test set uses option shuffling and answer-mapping randomization. In total, 2--3 annotators and one adjudicator label 2{,}298 stems over $\approx$1.5 months. Local audits find no missing images, duplicate released stems, or malformed P/R probes; details are in Appendix~\ref{app:annotation_details}--\ref{app:contamination}.

\paragraph{Scale, coverage, and validity properties.} The released result-bearing split contains 2{,}298 stems and 19{,}533 tasks: 2{,}298 F items, 2{,}298 R probes, and 14{,}937 P probes (avg $\approx$6.50 per stem). Table~\ref{tab:dataset_overview} gives the breakdown. Scale is secondary; the design favors deeper annotation and probe-level reuse. F and R are mostly medium difficulty (77.0\% and 89.7\%), while P is more dispersed (48.6\% easy, 33.8\% medium, 17.6\% hard). Four properties support diagnostic use: balanced RI/VP coverage, the $\approx$$6.5{:}1{:}1$ P/R/F ratio for within-stem pairing, frozen annotation audits (Appendix~\ref{app:iaa}), and contamination controls with hidden-test answer randomization.

\subsection{Evaluation protocol}
\label{sec:eval_protocol}

\paragraph{Evaluation levels and metrics.} \avrb\ reports L1 exact-match \sac{}, L2 StepAcc from a free-trace judge, and L3 AttrTag from first-failure stage crossed with perception-load tag. The main analyses use P/R/F accuracy, R--F gaps, stem-level failure decomposition, strict Binding Gap, L2 StepAcc, and \ssa{} gains. Auxiliary validity metrics include \stc{} (all P/R/F probes correct on a stem) and \vg{} ($\sac_{\text{full-image}}-\sac_{\text{text-only}}$); auxiliary audit summaries are in Appendix~\ref{app:aux_metrics}.

\paragraph{Free-trace L2 and \ssa.} Models only wrap the final choice as \verb|<ANSWER>X</ANSWER>|; the rest is a free trace. An external judge aligns the trace to S1--S4 ground truth and emits stage flags plus first-failure stage (Appendix~\ref{app:judge_calibration}). \ssa\ supplies verified stage content cumulatively from H0 raw input through H4 target-instance description; no condition includes the answer letter or downstream content. We report raw deltas, isotonic gains, and an irrelevant-S1 control.

\section{Experiments}
\label{sec:exp}

\subsection{Experimental setup}
\label{sec:exp_setup}

Our 24-row direct-mode pool includes five state-of-the-art proprietary frontier models\footnote{For proprietary models without archival reports, we cite the closest official public source at submission time (system card, model card, or release post).} (GPT-5.4~\citep{openai_gpt54_system_card}, Gemini-3.1-Pro~\citep{gemini31_model_card}, Claude-Opus-4.7~\citep{anthropic_opus47_model_report}, Qwen3.5-Plus~\citep{qwen35blog}, and grok-4.2-beta) and two open-source standalone frontier models (Kimi-K2.5~\citep{kimi25} and GLM-4.5V~\citep{glm45v}). Three open-source scaling families anchor the family-level analysis: Qwen3.5~\citep{qwen35blog} (0.8B, 2B, 4B, 9B, 27B, 35B-A3B, and 122B-A10B), InternVL3.5~\citep{internvl35} (1B, 2B, 4B, 8B, 14B, and 30B-A3B), and Gemma~4~\citep{gemma4_model_card} (E2B-it, E4B-it, 26B-A4B-it, and 31B-it). Headline family rows use direct mode; available thinking variants are reported only as paired $\Delta$ diagnostics and excluded from family scaling. The completed stage-wise diagnostic covers all seven Qwen3.5 rows on the full 2{,}298-item F split, with L2 StepAcc judged by a deterministic GPT-4o judge~\citep{openai_gpt4o_system_card} and \ssa{} replicated on Gemma~4 in Appendix~\ref{app:gemma4_ssa}. All runs use temperature 0, fixed max-token budgets, English stems, and full-image input. Metrics are defined in \S\ref{sec:eval_protocol}; the full pool with release dates is in Appendix~\ref{app:full_leaderboard}.

\subsection{Main leaderboard}
\label{sec:exp_main}

\begin{figure}[h]
  \centering
  \includegraphics[width=\textwidth]{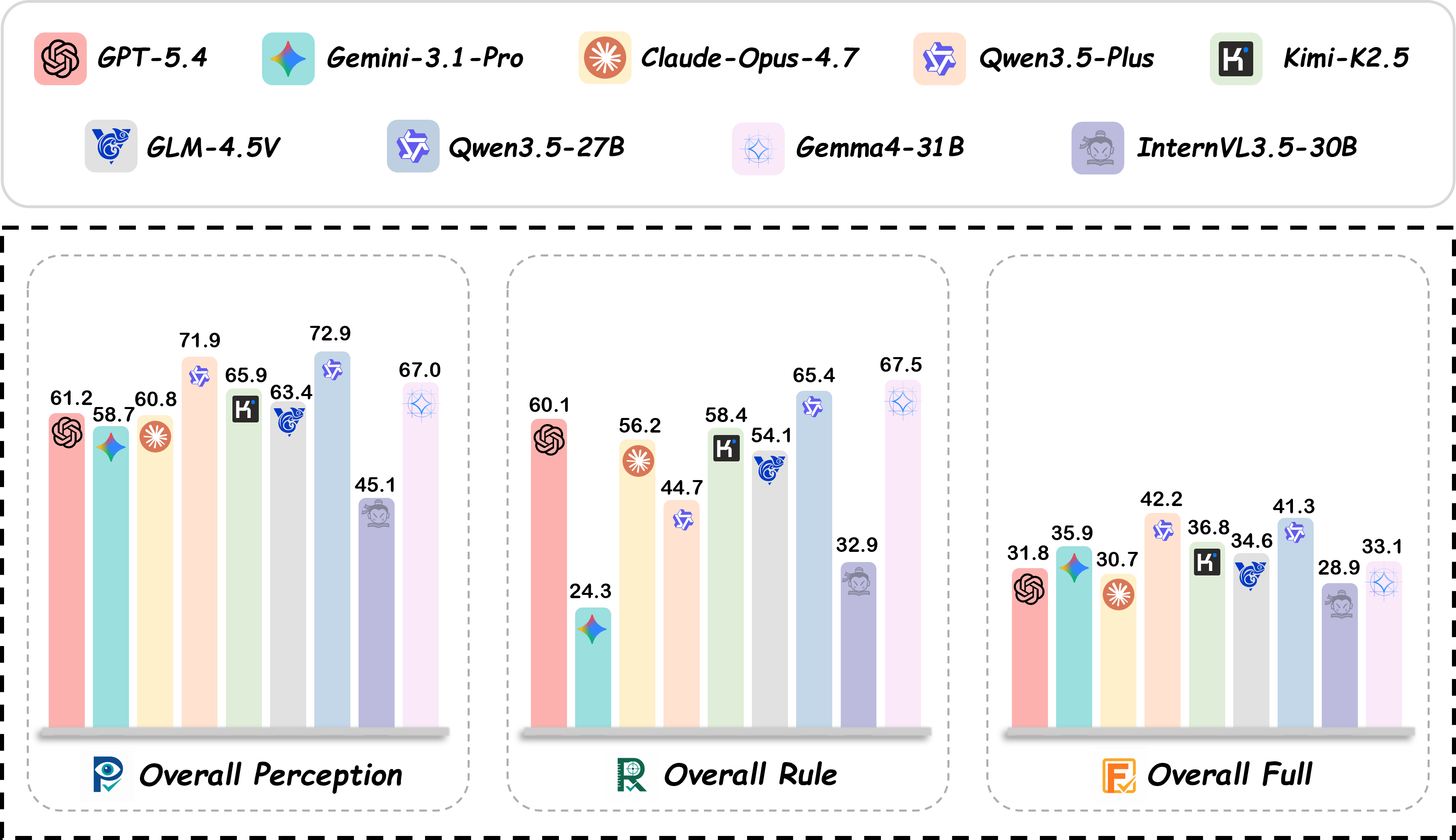}
  \caption{Aggregate P/R/F performance. Many models preserve stronger P or R accuracy while dropping on F.}
  \label{fig:model_performance}
\end{figure}

\paragraph{Three insights from the leaderboard.} Figure~\ref{fig:model_performance} summarizes the aggregate P/R/F profile behind Table~\ref{tab:main_leaderboard}. Full-item accuracy stays low in absolute terms: the best full-set open row, Qwen3.5-27B, only reaches $F{=}41.3\%$, and the five proprietary rows cluster between $28\%$ and $42\%$ F without dominating both R and F. Standalone open-source frontier rows sit in the same band (GLM-4.5V $34.6\%$ F, Kimi-K2.5 $36.8\%$ F), so the pattern is not tied to release type. Many rows also preserve higher P or R than F. Final-answer accuracy ranks models but does not identify the source of failure; \S\ref{sec:exp_chasm}--\S\ref{sec:exp_further} diagnose where it arises.

\definecolor{pairblue}{RGB}{232,240,255}
\definecolor{pairblue2}{RGB}{243,247,255}
\definecolor{secgray}{gray}{0.92}
\newcommand{\triplethead}{{\scriptsize\textcolor{black!75}{\textbf{P}}} & {\scriptsize\textcolor{black!75}{\textbf{R}}} & {\scriptsize\textcolor{black!75}{\textbf{F}}}}
\newcommand{\grouphead}[1]{\rowcolor{secgray}\multicolumn{31}{@{}c@{}}{\textbf{#1}}\\}
\newcommand{\thinklabel}{\hspace{0.9em}\textcolor{black!70}{$\hookrightarrow$ \textit{thinking} $\Delta$}}
\newcommand{\dpos}[1]{\textcolor{green!45!black}{+#1}}
\newcommand{\dneg}[1]{\textcolor{red!70!black}{#1}}

\begin{table}[H]
\centering
\caption{%
  \textbf{\avrb{} main leaderboard.}
  P/R/F accuracy (\%) for perception, rule, and full-item probes across all operations and RI/VP types.
  Thinking-$\Delta$ rows show thinking minus direct accuracy.
}
\label{tab:main_leaderboard}
\vspace{0.35ex}
\scriptsize
\setlength{\tabcolsep}{2.7pt}
\renewcommand{\arraystretch}{1.08}
\resizebox{\textwidth}{!}{%
\begin{tabular}{@{}>{\raggedright\arraybackslash}p{3.35cm} *{30}{c}@{}}
\toprule
\textbf{Model} &
\multicolumn{3}{c}{\textbf{Overall}} &
\multicolumn{15}{c}{\textbf{Rule Induction (RI)}} &
\multicolumn{12}{c}{\textbf{Visual Processing (VP)}} \\
\cmidrule(lr){2-4}\cmidrule(lr){5-19}\cmidrule(lr){20-31}
&
\multicolumn{3}{c}{\textbf{All}} &
\multicolumn{3}{c}{\textbf{Pos}} &
\multicolumn{3}{c}{\textbf{Sty}} &
\multicolumn{3}{c}{\textbf{Attr}} &
\multicolumn{3}{c}{\textbf{Qty}} &
\multicolumn{3}{c}{\textbf{Rel}} &
\multicolumn{3}{c}{\textbf{Fold}} &
\multicolumn{3}{c}{\textbf{View}} &
\multicolumn{3}{c}{\textbf{Rot}} &
\multicolumn{3}{c}{\textbf{Clos}} \\
& \triplethead & \triplethead & \triplethead & \triplethead &
\triplethead & \triplethead & \triplethead & \triplethead &
\triplethead & \triplethead \\
\midrule
\grouphead{Closed-source frontier}
\rowcolor{pairblue}
\micon{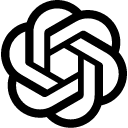}GPT-5.4             & 61.2 & 60.1 & 31.8 & 52.9 & 39.4 & 26.1 & 54.0 & 71.4 & 42.9 & 66.4 & 37.1 & 34.3 & 60.8 & 43.8 & 35.1 & 64.1 & 57.8 & 40.0 & 57.8 & 90.0 & 28.6 & 69.3 & 73.1 & 19.2 & 59.4 & 41.9 & 32.6 & 68.0 & 69.7 & 27.3 \\
\rowcolor{pairblue2}
\thinklabel                            & \dpos{2.4} & \dneg{-1.8} & \dneg{-4.6} & \dpos{3.1} & \dneg{-1.2} & \dneg{-4.2} & \dpos{2.0} & \dpos{0.5} & \dneg{-3.8} & \dpos{3.8} & \dneg{-2.5} & \dneg{-5.6} & \dpos{1.5} & \dneg{-1.0} & \dneg{-4.8} & \dpos{2.7} & \dneg{-0.8} & \dneg{-5.1} & \dpos{1.1} & \dneg{-3.2} & \dneg{-6.5} & \dpos{2.2} & \dneg{-2.4} & \dneg{-5.8} & \dpos{1.8} & \dneg{-1.9} & \dneg{-4.7} & \dpos{3.4} & \dneg{-1.6} & \dneg{-5.2} \\
\rowcolor{pairblue}
\micon{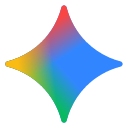}Gemini-3.1-Pro & 58.7 & 24.3 & 35.9 & 43.9 & 13.3 & 30.0 & 57.9 & 28.6 & 33.3 & 70.3 & 9.4 & 37.5 & 57.0 & 20.0 & 37.8 & 66.3 & 15.6 & 33.3 & 55.5 & 44.3 & 35.7 & 56.0 & 19.2 & 30.8 & 58.1 & 20.9 & 44.2 & 63.2 & 30.3 & 36.4 \\
\rowcolor{pairblue2}
\thinklabel                            & \dpos{1.7} & \dneg{-2.9} & \dneg{-4.8} & \dpos{2.4} & \dneg{-1.5} & \dneg{-4.2} & \dpos{1.5} & \dneg{-2.4} & \dneg{-5.0} & \dpos{2.8} & \dneg{-3.1} & \dneg{-5.8} & \dpos{1.0} & \dneg{-2.0} & \dneg{-4.5} & \dpos{1.9} & \dneg{-3.3} & \dneg{-5.2} & \dpos{0.8} & \dneg{-4.5} & \dneg{-6.0} & \dpos{1.6} & \dneg{-2.7} & \dneg{-5.1} & \dpos{1.2} & \dneg{-2.5} & \dneg{-4.6} & \dpos{2.3} & \dneg{-3.8} & \dneg{-5.4} \\
\rowcolor{pairblue}
\micon{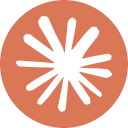}Claude-Opus-4.7          & 60.8 & 56.2 & 30.7 & 52.0 & 39.4 & 26.1 & 54.7 & 53.3 & 32.8 & 66.0 & 35.7 & 35.2 & 64.8 & 49.2 & 34.1 & 67.9 & 48.1 & 33.7 & 55.8 & 83.7 & 25.8 & 64.8 & 75.7 & 34.7 & 56.9 & 42.1 & 28.3 & 66.2 & 58.6 & 30.9 \\
\rowcolor{pairblue2}
\thinklabel                            & \dpos{2.8} & \dneg{-1.4} & \dneg{-4.0} & \dpos{3.0} & \dneg{-1.0} & \dneg{-3.5} & \dpos{2.1} & \dpos{0.2} & \dneg{-3.2} & \dpos{3.5} & \dneg{-2.0} & \dneg{-5.1} & \dpos{1.8} & \dneg{-0.7} & \dneg{-4.1} & \dpos{2.6} & \dneg{-1.2} & \dneg{-4.8} & \dpos{1.2} & \dneg{-3.0} & \dneg{-6.2} & \dpos{2.0} & \dneg{-2.1} & \dneg{-4.9} & \dpos{1.7} & \dneg{-1.8} & \dneg{-4.4} & \dpos{3.0} & \dneg{-1.5} & \dneg{-5.0} \\
\rowcolor{pairblue}
\micon{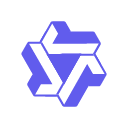}Qwen3.5-Plus       & 71.9 & 44.7 & 42.2 & 57.1 & 27.3 & 32.0 & 74.4 & 38.5 & 38.5 & 81.1 & 44.0 & 48.0 & 69.8 & 41.9 & 42.4 & 81.4 & 52.9 & 50.0 & 71.4 & 63.6 & 45.5 & 66.7 & 85.7 & 71.4 & 62.4 & 44.4 & 42.9 & 77.2 & 37.9 & 28.6 \\
\rowcolor{pairblue2}
\thinklabel                            & \dpos{1.6} & \dneg{-3.8} & \dneg{-6.8} & \dpos{2.4} & \dneg{-2.5} & \dneg{-5.8} & \dpos{1.2} & \dneg{-3.0} & \dneg{-6.5} & \dpos{2.8} & \dneg{-4.6} & \dneg{-7.4} & \dpos{0.9} & \dneg{-3.8} & \dneg{-6.9} & \dpos{1.5} & \dneg{-4.2} & \dneg{-7.1} & \dpos{0.5} & \dneg{-6.5} & \dneg{-8.2} & \dpos{1.0} & \dneg{-5.1} & \dneg{-7.8} & \dpos{0.8} & \dneg{-3.7} & \dneg{-6.6} & \dpos{2.0} & \dneg{-4.8} & \dneg{-7.5} \\
\rowcolor{pairblue}
\micon{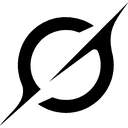}grok-4.2-beta        & 49.3 & 31.3 & 28.1 & 46.7 & 30.0 & 13.3 & 45.2 & 38.1 & 47.6 & 55.2 & 6.2 & 28.1 & 41.5 & 13.3 & 26.7 & 48.1 & 31.1 & 33.3 & 53.5 & 47.1 & 32.9 & 49.5 & 34.6 & 19.2 & 46.5 & 32.6 & 27.9 & 53.7 & 39.4 & 21.2 \\
\rowcolor{pairblue2}
\thinklabel                            & \dpos{2.1} & \dneg{-1.1} & \dneg{-3.9} & \dpos{2.6} & \dneg{-0.8} & \dneg{-3.2} & \dpos{1.9} & \dneg{-0.3} & \dneg{-3.5} & \dpos{3.0} & \dneg{-1.7} & \dneg{-4.7} & \dpos{1.2} & \dneg{-1.0} & \dneg{-3.8} & \dpos{2.2} & \dneg{-1.2} & \dneg{-4.0} & \dpos{0.9} & \dneg{-2.5} & \dneg{-5.3} & \dpos{1.7} & \dneg{-1.8} & \dneg{-4.5} & \dpos{1.3} & \dneg{-1.5} & \dneg{-3.7} & \dpos{2.4} & \dneg{-1.3} & \dneg{-4.4} \\
\midrule
\grouphead{Open-source frontier models}
\micon{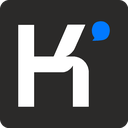}Kimi-K2.5         & 65.9 & 58.4 & 36.8 & 55.0 & 47.5 & 33.3 & 66.7 & 63.2 & 42.9 & 71.9 & 62.5 & 53.1 & 58.9 & 49.2 & 31.1 & 68.1 & 64.1 & 33.3 & 68.2 & 74.0 & 37.1 & 65.4 & 76.5 & 46.2 & 63.1 & 54.0 & 27.9 & 73.6 & 65.2 & 36.4 \\
\micon{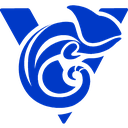}GLM-4.5V   & 63.4 & 54.1 & 34.6 & 54.8 & 39.5 & 31.2 & 61.6 & 51.8 & 33.8 & 68.5 & 44.2 & 38.5 & 64.7 & 50.4 & 36.4 & 66.2 & 56.8 & 35.1 & 60.1 & 70.5 & 32.4 & 67.4 & 69.2 & 40.1 & 62.0 & 49.5 & 31.8 & 65.5 & 55.7 & 33.9 \\
\midrule
\grouphead{Open-source: Qwen3.5 family}
\micon{Qwen3}Qwen3.5-122B-A10B & 55.7 & 29.6 & 33.8 & 61.4 & 30.0 & 38.9 & 50.4 & 20.4 & 32.8 & 61.6 & 31.0 & 44.6 & 64.1 & 39.1 & 37.5 & 58.5 & 28.6 & 32.3 & 36.4 & 9.2 & 24.7 & 56.1 & 22.0 & 37.0 & 59.2 & 34.5 & 32.8 & 72.2 & 65.0 & 34.5 \\
\micon{Qwen3}Qwen3.5-35B-A3B & 72.6 & 59.8 & 39.3 & 67.5 & 34.0 & 32.5 & 67.4 & 35.8 & 40.1 & 75.2 & 57.7 & 45.1 & 74.7 & 53.8 & 43.8 & 77.3 & 51.5 & 43.4 & 69.7 & 76.4 & 34.1 & 75.6 & 78.6 & 48.0 & 71.3 & 53.1 & 38.6 & 75.2 & 79.1 & 33.2 \\
\micon{Qwen3}Qwen3.5-27B & 72.9 & 65.4 & 41.3 & 69.5 & 43.3 & 35.0 & 69.6 & 50.4 & 39.4 & 76.6 & 55.9 & 46.0 & 74.7 & 58.9 & 43.8 & 76.1 & 60.3 & 42.4 & 70.0 & 83.0 & 38.0 & 75.1 & 82.7 & 49.7 & 72.3 & 59.3 & 39.3 & 73.3 & 77.7 & 41.8 \\
\thinklabel & \dpos{1.4} & \dneg{-4.8} & \dneg{-7.2} & \dpos{1.8} & \dneg{-3.5} & \dneg{-6.5} & \dpos{1.2} & \dneg{-4.0} & \dneg{-7.1} & \dpos{2.0} & \dneg{-5.2} & \dneg{-8.0} & \dpos{0.9} & \dneg{-4.6} & \dneg{-7.3} & \dpos{1.5} & \dneg{-5.0} & \dneg{-7.5} & \dpos{0.5} & \dneg{-7.8} & \dneg{-9.0} & \dpos{1.1} & \dneg{-6.2} & \dneg{-8.4} & \dpos{0.7} & \dneg{-4.7} & \dneg{-6.8} & \dpos{1.9} & \dneg{-5.5} & \dneg{-8.1} \\
\micon{Qwen3}Qwen3.5-9B & 68.9 & 60.1 & 37.4 & 63.2 & 33.5 & 31.0 & 66.5 & 40.1 & 32.1 & 71.3 & 42.7 & 40.8 & 68.1 & 50.8 & 44.5 & 73.0 & 55.9 & 40.1 & 65.8 & 84.1 & 34.5 & 74.0 & 79.2 & 48.6 & 68.1 & 58.6 & 32.8 & 72.7 & 67.7 & 33.2 \\
\thinklabel & \dpos{1.1} & \dneg{-5.5} & \dneg{-7.8} & \dpos{1.6} & \dneg{-4.2} & \dneg{-7.0} & \dpos{0.8} & \dneg{-4.8} & \dneg{-7.5} & \dpos{1.8} & \dneg{-5.7} & \dneg{-8.3} & \dpos{0.6} & \dneg{-5.2} & \dneg{-7.9} & \dpos{1.2} & \dneg{-5.9} & \dneg{-8.0} & \dpos{0.3} & \dneg{-8.5} & \dneg{-9.6} & \dpos{0.9} & \dneg{-6.8} & \dneg{-8.8} & \dpos{0.5} & \dneg{-5.0} & \dneg{-7.2} & \dpos{1.5} & \dneg{-6.0} & \dneg{-8.4} \\
\micon{Qwen3}Qwen3.5-4B & 66.5 & 56.7 & 33.8 & 62.5 & 37.4 & 27.1 & 61.6 & 39.4 & 26.3 & 69.9 & 53.1 & 36.2 & 67.0 & 51.8 & 39.8 & 69.9 & 51.5 & 34.7 & 64.3 & 68.9 & 31.1 & 69.9 & 75.1 & 43.4 & 64.0 & 50.7 & 33.8 & 70.7 & 70.0 & 31.4 \\
\thinklabel & \dpos{0.8} & \dneg{-4.9} & \dneg{-8.4} & \dpos{1.2} & \dneg{-3.8} & \dneg{-7.5} & \dpos{0.5} & \dneg{-4.2} & \dneg{-8.0} & \dpos{1.5} & \dneg{-5.3} & \dneg{-8.9} & \dpos{0.4} & \dneg{-4.7} & \dneg{-8.5} & \dpos{0.9} & \dneg{-5.5} & \dneg{-8.8} & \dpos{0.2} & \dneg{-7.0} & \dneg{-9.8} & \dpos{0.6} & \dneg{-6.1} & \dneg{-9.0} & \dpos{0.3} & \dneg{-4.6} & \dneg{-7.8} & \dpos{1.1} & \dneg{-5.4} & \dneg{-8.7} \\
\micon{Qwen3}Qwen3.5-2B & 61.4 & 60.5 & 35.9 & 52.6 & 32.5 & 28.1 & 55.6 & 45.3 & 40.1 & 62.7 & 51.6 & 41.8 & 61.6 & 45.2 & 35.8 & 63.1 & 56.6 & 32.3 & 61.6 & 84.1 & 34.8 & 65.2 & 82.7 & 46.8 & 60.2 & 55.9 & 34.1 & 66.3 & 69.1 & 35.5 \\
\thinklabel & \dpos{0.2} & \dneg{-5.5} & \dneg{-8.7} & \dpos{0.8} & \dneg{-4.3} & \dneg{-7.9} & \dneg{-0.1} & \dneg{-4.8} & \dneg{-8.3} & \dpos{1.0} & \dneg{-5.9} & \dneg{-9.2} & \dneg{-0.2} & \dneg{-5.2} & \dneg{-8.7} & \dpos{0.3} & \dneg{-5.8} & \dneg{-9.0} & \dneg{-0.7} & \dneg{-7.4} & \dneg{-10.1} & \dpos{0.0} & \dneg{-6.6} & \dneg{-9.4} & \dneg{-0.3} & \dneg{-5.2} & \dneg{-8.2} & \dpos{0.7} & \dneg{-6.0} & \dneg{-9.0} \\
\micon{Qwen3}Qwen3.5-0.8B & 41.5 & 48.7 & 30.2 & 32.5 & 26.6 & 25.6 & 38.0 & 31.4 & 24.8 & 45.9 & 44.6 & 34.7 & 43.6 & 37.1 & 26.4 & 43.2 & 37.0 & 33.0 & 38.6 & 70.4 & 32.4 & 46.2 & 67.6 & 32.9 & 39.3 & 45.5 & 31.7 & 47.9 & 58.6 & 25.9 \\
\thinklabel & \dneg{-0.6} & \dneg{-6.2} & \dneg{-9.1} & \dpos{0.2} & \dneg{-5.0} & \dneg{-8.4} & \dneg{-0.8} & \dneg{-5.5} & \dneg{-8.8} & \dpos{0.3} & \dneg{-6.6} & \dneg{-9.5} & \dneg{-1.0} & \dneg{-5.8} & \dneg{-9.0} & \dneg{-0.4} & \dneg{-6.3} & \dneg{-9.2} & \dneg{-1.8} & \dneg{-8.0} & \dneg{-10.5} & \dneg{-0.7} & \dneg{-7.2} & \dneg{-9.9} & \dneg{-1.2} & \dneg{-6.0} & \dneg{-8.7} & \dpos{0.1} & \dneg{-6.8} & \dneg{-9.4} \\
\midrule
\grouphead{Open-source: InternVL3.5 family}
\micon{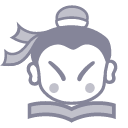}InternVL3.5-30B-A3B & 45.1 & 32.9 & 28.9 & 36.1 & 16.3 & 29.1 & 40.1 & 21.9 & 32.1 & 47.0 & 13.1 & 28.6 & 43.3 & 27.1 & 27.8 & 47.2 & 27.9 & 32.7 & 44.0 & 62.0 & 26.6 & 50.5 & 36.4 & 35.3 & 43.3 & 30.3 & 25.9 & 53.0 & 27.3 & 27.3 \\
\micon{internlm-color}InternVL3.5-14B & 43.2 & 30.7 & 28.1 & 34.8 & 15.9 & 27.4 & 38.6 & 23.4 & 30.9 & 46.1 & 12.0 & 27.6 & 42.0 & 29.8 & 27.3 & 45.5 & 25.7 & 31.2 & 41.8 & 58.5 & 25.9 & 49.0 & 35.1 & 34.4 & 41.2 & 28.8 & 25.1 & 51.7 & 25.9 & 26.5 \\
\micon{internlm-color}InternVL3.5-8B  & 39.5 & 33.6 & 26.7 & 30.2 & 18.1 & 25.8 & 36.9 & 25.6 & 29.7 & 42.8 & 14.8 & 25.2 & 37.4 & 31.5 & 26.4 & 43.1 & 28.5 & 29.0 & 39.6 & 60.4 & 23.7 & 47.2 & 37.9 & 32.8 & 38.0 & 31.6 & 23.9 & 49.6 & 28.4 & 24.0 \\
\micon{internlm-color}InternVL3.5-4B  & 36.8 & 27.9 & 24.8 & 28.7 & 13.6 & 24.2 & 32.4 & 20.7 & 27.8 & 39.9 & 10.4 & 23.5 & 34.1 & 25.0 & 24.9 & 38.2 & 22.1 & 27.0 & 35.3 & 52.8 & 22.6 & 42.6 & 30.2 & 29.1 & 35.7 & 27.4 & 22.2 & 45.4 & 23.1 & 23.6 \\
\micon{internlm-color}InternVL3.5-2B  & 34.9 & 29.4 & 23.1 & 26.1 & 15.4 & 22.0 & 31.2 & 18.9 & 25.5 & 37.5 & 12.2 & 22.9 & 32.7 & 27.8 & 23.0 & 36.0 & 21.5 & 25.8 & 33.1 & 55.3 & 21.4 & 39.8 & 32.6 & 27.0 & 32.9 & 29.1 & 21.1 & 43.5 & 24.9 & 22.5 \\
\micon{internlm-color}InternVL3.5-1B  & 30.6 & 24.8 & 20.7 & 23.9 & 11.7 & 20.1 & 28.1 & 17.4 & 23.2 & 33.8 & 8.9 & 20.6 & 29.5 & 21.2 & 21.4 & 31.4 & 19.3 & 23.5 & 30.2 & 48.7 & 19.8 & 36.6 & 27.5 & 24.8 & 29.6 & 24.0 & 19.7 & 39.0 & 20.1 & 20.9 \\
\midrule
\grouphead{Open-source: Gemma 4 family}
\micon{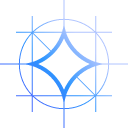}Gemma-4-31B-it & 67.0 & 67.5 & 33.1 & 51.2 & 58.6 & 28.6 & 64.2 & 62.7 & 27.0 & 47.4 & 72.4 & 36.2 & 57.9 & 69.4 & 40.8 & 67.0 & 71.4 & 34.7 & 86.1 & 64.3 & 31.3 & 86.1 & 71.8 & 38.7 & 57.6 & 65.9 & 29.0 & 71.4 & 71.7 & 30.0 \\
\micon{gemma-color}Gemma-4-26B-A4B-it & 71.7 & 65.4 & 33.0 & 60.6 & 58.9 & 28.1 & 58.4 & 61.8 & 32.8 & 59.6 & 70.0 & 39.4 & 67.6 & 64.8 & 36.8 & 70.4 & 69.0 & 31.0 & 88.8 & 63.3 & 32.0 & 82.1 & 68.6 & 38.2 & 63.4 & 64.4 & 30.3 & 75.5 & 68.6 & 30.9 \\
\micon{gemma-color}Gemma-4-E4B-it & 66.0 & 55.0 & 30.6 & 46.3 & 52.1 & 28.1 & 51.8 & 53.2 & 26.3 & 61.0 & 56.6 & 39.4 & 67.2 & 50.2 & 32.8 & 66.0 & 56.3 & 24.9 & 83.3 & 55.5 & 30.0 & 80.3 & 59.9 & 38.7 & 54.1 & 53.6 & 24.5 & 64.1 & 58.2 & 35.0 \\
\micon{gemma-color}Gemma-4-E2B-it & 58.9 & 43.0 & 28.6 & 38.9 & 40.9 & 21.2 & 35.8 & 42.1 & 24.8 & 53.5 & 38.8 & 35.2 & 52.5 & 38.3 & 30.8 & 46.5 & 35.7 & 25.6 & 82.4 & 47.6 & 29.2 & 83.2 & 49.1 & 36.4 & 48.3 & 43.2 & 24.1 & 67.3 & 47.3 & 30.9 \\
\bottomrule
\end{tabular}
}%
\end{table}

\subsection{The R--F chasm and binding-dominant failures}
\label{sec:exp_chasm}

\begin{findingbox}
\textbf{Finding~1.} Final-answer accuracy ranks the 24 direct rows cleanly, \textbf{but} R outperforms F on 22 of them; on the 20 stem-paired rows ``can't bind'' is the single largest F-wrong category on 11 of 20 rows, and a strict Binding Gap of $0.51$ persists even when both P and R are correct on the same stem. The chasm is a behavioral rule-to-instance binding bottleneck, not a missing-rule pattern.
\end{findingbox}

\begin{figure}[t]
  \centering
  \includegraphics[width=0.92\textwidth]{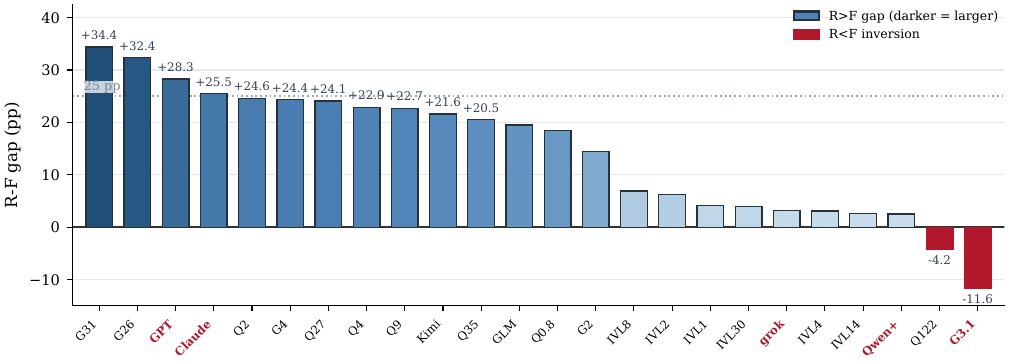}
  \vspace{-1.0ex}
  \includegraphics[width=0.92\textwidth]{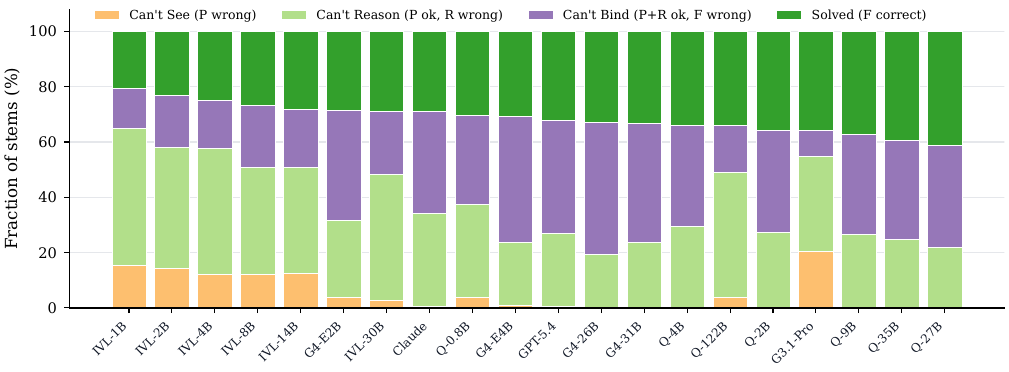}
  \vspace{-1.0ex}
  \includegraphics[width=0.92\textwidth]{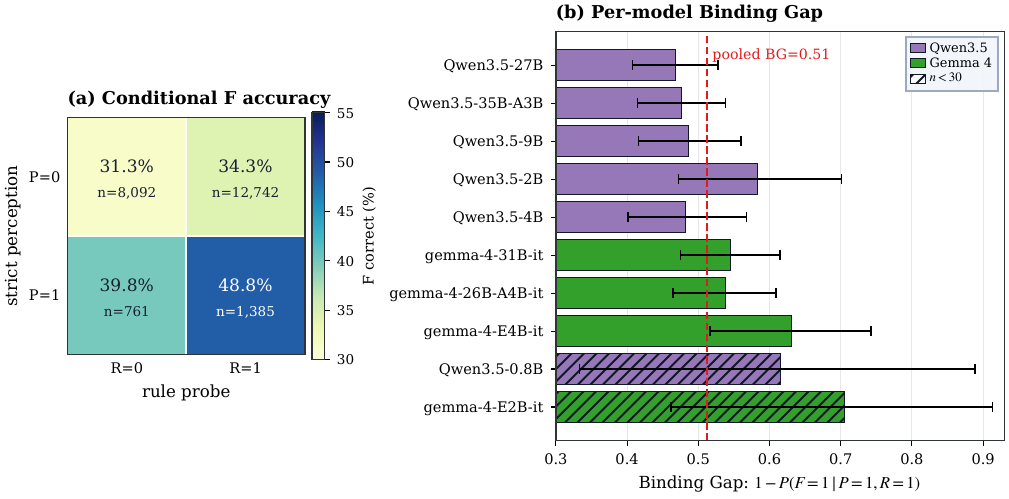}
  \caption{Binding evidence. \textbf{Top:} R--F chasm. \textbf{Middle:} failure decomposition. \textbf{Bottom:} strict conditional Binding Gap.}
  \label{fig:binding_evidence}
  \label{fig:rf_chasm}
  \label{fig:failure_decomp}
  \label{fig:binding_gap_synergy}
\end{figure}

We give three layers of evidence: \textbf{(i)} the aggregate R--F chasm across all 24 direct rows; \textbf{(ii)} a stem-level failure decomposition on the 20 rows with paired P/R/F; and \textbf{(iii)} a strict conditional Binding Gap controlling for joint P/R correctness (Fig.~\ref{fig:binding_evidence}).

\paragraph{(i) Aggregate R--F chasm.} The R--F gap is positive on 22 of 24 direct rows. On low-R rows it shrinks to a few points, but on the higher-R rows it widens to 18--34\,pp; representative cases include Qwen3.5-27B ($\Delta\!=\!24.1$\,pp), Gemma-4-31B-it ($\Delta\!=\!34.4$\,pp), Gemma-4-26B-A4B-it ($\Delta\!=\!32.4$\,pp), and GPT-5.4 ($\Delta\!=\!28.3$\,pp). The two exceptions are rule-probe-collapse rows rather than high-R binding successes: Gemini-3.1-Pro ($R{=}24.3\%$, $F{=}35.9\%$) and Qwen3.5-122B-A10B ($R{=}29.6\%$, $F{=}33.8\%$) both fall to or below chance on R. \textbf{Finding~1a.}~R systematically exceeds F across the leaderboard; the only inversions are rule-collapse rows, not high-R binding successes.

\paragraph{(ii) Stem-level failure decomposition.} For each stem, a model's joint P/R/F outcome partitions into four exhaustive categories: ``solved'' (F correct), ``can't bind'' (at least one P probe and the R probe correct, F wrong), ``can't reason'' (at least one P probe correct, R wrong, F wrong), and ``can't see'' (no P probe correct, F wrong). The strict analysis in (iii) requires all P probes instead. The breakdown uses only L1 outcomes and covers 20 of 24 direct rows. ``Can't bind'' is the largest failure category on 11 of 20 rows. Across stronger Qwen3.5/Gemma~4 rows it sits in the mid-30s to high-40s of stems (e.g., Qwen3.5-27B $36.7\%$, Qwen3.5-35B-A3B $35.7\%$, Gemma-4-26B-A4B-it $47.5\%$, and Gemma-4-31B-it $43.3\%$); conditionally, among F-wrong stems on these rows, about 56--71\% have R correct. The proprietary leaders echo the pattern (GPT-5.4 $40.9\%$, Claude-Opus-4.7 $36.7\%$ can't-bind). Gemini-3.1-Pro is the non-MoE outlier (34.5\% can't-reason, 20.4\% can't-see, only 9.2\% can't-bind), corroborating its rule-collapse profile above. \textbf{Finding~1b.}~On stronger models, F errors are dominated by stems where perception and rule are both correct; binding is the dominant failure category exposed by the shared-stem design.

\paragraph{(iii) Strict conditional Binding Gap.} To recast the four-quadrant view as a conditional probability, we compute strict stem-level perception correctness ($P{=}1$ only when every annotated P probe for the stem is correct, $6.50$ per stem on average) and report $P(F{=}1 \mid P{=}1, R{=}1)$. On the open-source full-split rows from Qwen3.5 and Gemma~4, the jointly successful stratum still solves F only $48.8\%$ of the time ($n{=}1{,}385$), giving a pooled Binding Gap $\mathrm{BG} = 1 - P(F{=}1 \mid P{=}1, R{=}1) = 0.512$. The strongest full-split row, Qwen3.5-27B, retains $\mathrm{BG}=0.467$ with bootstrap 95\% CI $[0.407, 0.528]$. A model-fixed-effect logistic check $F \sim P + R + P{:}R + \alpha_m$ gives a positive but small interaction ($\hat{\gamma}=0.25$, $z=2.37$, $p=0.018$) that adds only $0.0002$ McFadden pseudo-$R^2$ over the P/R main effects: joint P/R correctness helps but does not erase the Binding Gap. The gap also holds under a stricter derivation-checked R criterion, where the stated rule must match the gold S2 annotation: F is still wrong $48\%$ of the time (95\% CI $[0.45, 0.51]$, $n{=}1{,}100$), so it is not driven by R answers reached through elimination (Appendix~\ref{app:binding_gap_ci}). \textbf{Finding~1c.}~Even after conditioning on a stem's P\,and\,R both being correct, the strongest open rows still miss F more than $40\%$ of the time.

\paragraph{Boundary.} The interpretation is behavioral, not mechanistic: ``can't bind'' depends on the shared-stem design, and the positive P/R interaction rules out a mechanistic ``synergy deficit'' reading. $P{=}1$ means that the model passed all annotated perception probes. The chasm is also operation-dependent: VP-Fold is especially large for several rows, while Gemma~4 keeps large gaps across both RI and VP and Gemini-3.1-Pro/Qwen3.5-122B-A10B remain rule-collapse cases. Per-operation tables are in Appendix~\ref{app:full_leaderboard}.

\subsection{S1--S4 and \ssa{} localization}
\label{sec:exp_stage}

\begin{findingbox}
\textbf{Finding~2.} The Qwen3.5 S1--S4/\ssa{} diagnostic points to S3 Map: S3 is the weakest L2 stage on every full-split row, the largest \ssa{} gain appears when S3 alignment is injected, and independent human labels reproduce the S3 attribution.
\end{findingbox}

\paragraph{L2 stage profile.} The L2 judge covers all 2{,}298 F items for seven Qwen3.5 rows. S3 Map is the weakest stage on every row, falling to mean StepAcc $30.5\%$, well below S1 ($43.5\%$), S2 ($42.2\%$), and S4 ($45.9\%$). Dense scaling improves all stages, but MoE variants do not outperform 27B; we treat first-failure histograms as secondary, since $23.8\%$--$26.9\%$ of traces carry inconsistent stage flags.

\paragraph{\ssa{} intervention profile.} The injection sweep concentrates the gain at S3. H1 alone does not help (H1$-$H0 is negative on every row, between $-9.5$ and $-0.5$\,pp; the irrelevant-S1 control is also below H0), the largest raw jump is H2$\rightarrow$H3 on every row (mean $+10.9$\,pp raw and $+14.5$\,pp isotonic), and H4 raises raw accuracy over H0 by $+10.1$ to $+26.3$\,pp. Gemma~4 shows a similar S2/S3-centered profile, with smaller variants peaking at H2$\rightarrow$H3 and the larger 26B-A4B and 31B variants peaking at H1$\rightarrow$H2 (Appendix~\ref{app:gemma4_ssa}).

\begin{figure}[t]
  \centering
  \includegraphics[width=0.82\textwidth]{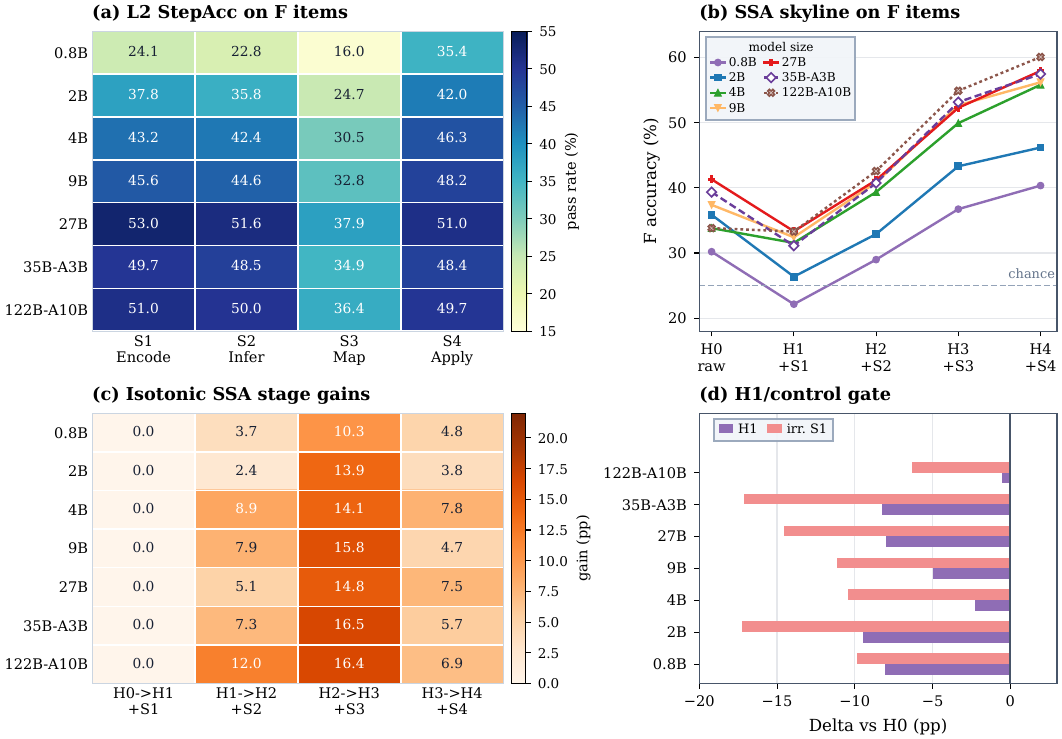}
  \caption{Qwen3.5 S1--S4/\ssa{} localization: S3 is weakest and gains concentrate at H2/H3.}
  \label{fig:stage_ssa_qwen35}
\end{figure}

\paragraph{Human validation.} Three independent annotators relabel S1--S4 on 180 stratified F traces. Judge--human agreement is substantial at every stage, and the human labels reproduce the S3 attribution: S3 remains the most frequent first-failure stage (45\% of F-wrong traces, against 48\% under the judge; Appendix~\ref{app:judge_calibration}).

\paragraph{Boundary.} This is behavioral localization with a controlled input intervention, not a mechanistic account~\citep{pearl2009,rubin1974}. The full L2 and \ssa{} evidence covers Qwen3.5 with a deterministic GPT-4o trace judge as the primary scorer ($n_\text{repeat}{=}1$), and a second judge (Qwen3.5-Plus) preserves the S3-weakest ordering on 6/7 Qwen3.5 rows with Spearman $\rho{=}0.93$ on the family ranking (Appendix~\ref{app:second_judge}); the Gemma~4 replication is appendix-only.

\subsection{Scaling and thinking do not close the gap}
\label{sec:exp_further}

\begin{findingbox}
\textbf{Finding~3.} Within-family scaling does not close the F-side binding gap. Qwen3.5 dense peaks at 27B and the 122B-A10B MoE variant collapses on R, while Gemma~4 and InternVL3.5 improve on F yet keep the R--F chasm.
\end{findingbox}

\paragraph{Family scaling.} Within-family scaling shifts P and R far more than F. Qwen3.5 improves through 27B ($F{=}41.3\%$, $R{=}65.4\%$), but the 122B-A10B MoE variant breaks the trend (R drops to $29.6\%$, with VP-Fold $R$ at $9.2\%$); Gemma~4 scales monotonically on F (from $28.6\%$ to $33.1\%$); and InternVL3.5 shows a noisier compact-to-large gain (F from $20.7\%$ to $28.9\%$). A larger model can raise P or R while F stays nearly flat, which a single leaderboard number cannot show.

\begin{figure}[t]
  \centering
  \includegraphics[width=0.90\textwidth]{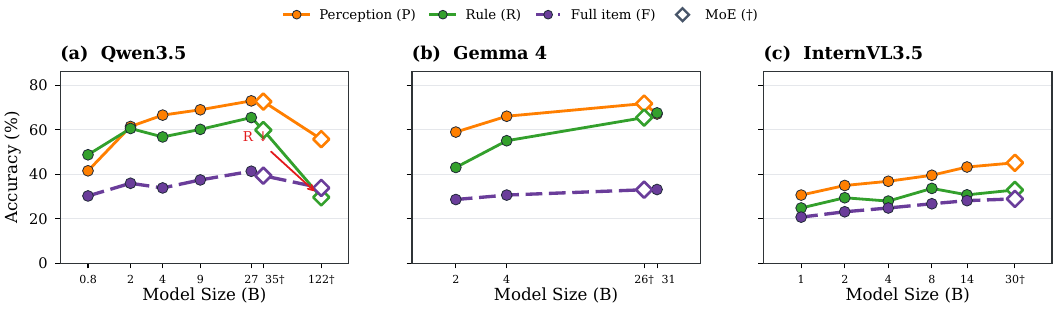}
  \caption{Family scaling: Qwen3.5 peaks pre-MoE; Gemma~4 improves on F; R--F gaps remain.}
  \label{fig:family_scaling}
\end{figure}

\begin{findingbox}
\textbf{Finding~4.} Explicit thinking does not repair the F-side binding gap. Across paired direct/thinking rows, P rises on nine of ten rows but R and F fall on every row, and the added F errors come mainly from rule-to-instance correspondence drift.
\end{findingbox}

\paragraph{Paired direct-vs-thinking deltas.} We report $\tgx = \sac_{\text{think},X} - \sac_{\text{non-think},X}$ for matched direct and thinking rows. The signed pattern is uniform: P rises on nine of ten rows; R and F fall on all ten. Full paired values plus a token-budget audit (Tab.~\ref{tab:token_budget}) ruling out truncation are in Appendix~\ref{app:thinkgain_scatter}.

\paragraph{What thinking breaks.} We label the first decisive error of each direct and thinking trace on 180 stems (20 per operation) for all ten pairs, blind to condition (Appendix~\ref{app:thinking_audit}). Correspondence drift, where the model states the correct rule but applies it to the wrong instance, rises the most (from 13.8 to 22.4 per 100 F items). Perception errors and rule-absent traces decrease. Thinking thus helps perception but loses the link between the rule and the target instance. Drift is also more frequent in longer traces: the longest third of thinking traces drifts twice as often as the shortest third (30.4 vs.\ 15.1 per 100 items).

\begin{figure}[t]
  \centering
  \includegraphics[width=0.90\textwidth]{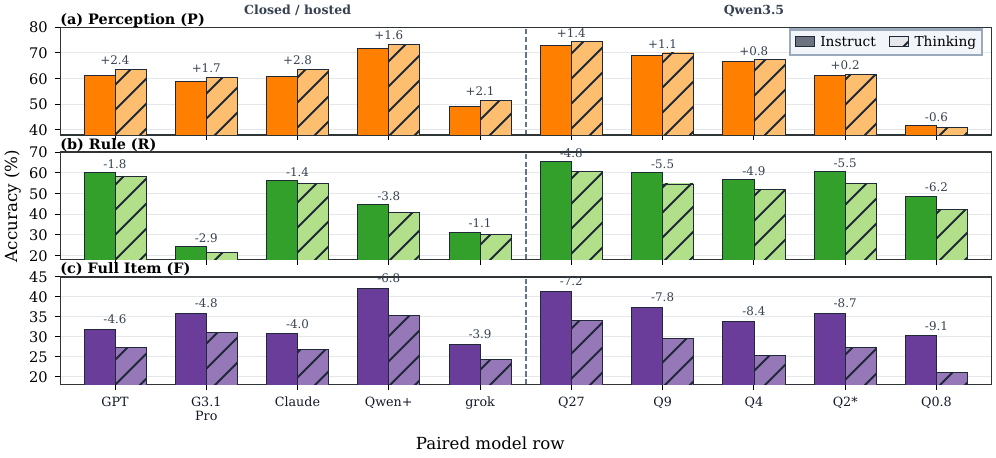}
  \caption{Paired direct versus thinking modes. Thinking lifts P on most rows but lowers R and F on all rows.}
  \label{fig:thinking_paired_bars_all}
\end{figure}

\vspace{-0.4ex}

\section{Conclusion, Limitations, and Future Work}
\label{sec:conclusion}
\label{sec:discussion}

\paragraph{Takeaway.} \avrb{} evaluates MLLMs with shared-stem perception, rule, and full-item probes, four-stage process annotations, and paired thinking controls. Rule accuracy exceeds full-item accuracy on 22 of 24 direct-mode rows, and a strict conditional Binding Gap of $0.51$ remains when perception and rule are both correct on the same stem. The Qwen3.5 stage and \ssa{} diagnostic places the dominant failure at S3 rule-to-instance mapping. AVR errors thus arise mainly from binding visual evidence to inferred rules, not from missing perception or rule induction, calling for leaderboards that locate where the solution path breaks, not only which model ranks higher.

\paragraph{Limitations.} \avrb{} uses provenance screening, dual-VLM pre-screening, and human adjudication across 2{,}298 stems, raising label reliability but limiting scale and non-English coverage. Stage and \ssa{} diagnostics target Qwen3.5 (Gemma~4 replicates in appendix); localization depends on a GPT-4o trace judge, with a second judge (Qwen3.5-Plus) replicating the S3 ranking (App.~\ref{app:second_judge}). Per-item S3 labels are noisier than at the other stages, for the judge and for human annotators alike, though the aggregate pattern holds. Findings are behavioral diagnostics, not causal claims about an internal binding module.

\paragraph{Future work.} Next steps include extending the stage and \ssa{} diagnostics to more model families and expanding \avrb{} to multilingual settings and richer reasoning-trace evaluations. These extensions would test whether perception-grounded training and vision-aware reasoning preserve rule-to-slot correspondence through final-answer selection. \ssa{} raises F most when verified S3 alignment is supplied ($+10.9$\,pp at H2$\rightarrow$H3), which makes S3 alignment a concrete target for training~\citep{he2026advantage} and prompting methods. Probing S3 at the attention and representation level is a natural next step toward a mechanistic account.

\begin{ack}
This work was supported by the National Natural Science Foundation of China under Grants 62306331 and 62407037, and by the CAAI Youth Talent Lifting Project under Grant CAAI2023-2025QNRC001.
We declare no competing interests.
\end{ack}

\clearpage
\bibliographystyle{plainnat}
\bibliography{references}

\clearpage
\appendix

\newcommand{\manualfigplaceholder}[4]{%
  \begin{figure}[H]
    \centering
    \begin{tcolorbox}[
      width=0.96\textwidth,
      colback=black!2,
      colframe=black!45,
      boxrule=0.5pt,
      arc=2pt,
      left=7pt,
      right=7pt,
      top=5pt,
      bottom=5pt,
      enhanced jigsaw
    ]
      {\small\textbf{Manual figure to draw:} \texttt{#1}}\par\vspace{0.4ex}
      {\footnotesize #2}
    \end{tcolorbox}
    \caption{#3}
    \label{#4}
  \end{figure}
}

\makeatletter
\newcommand{\appcontentssection}[3]{%
  \par\addvspace{0.95ex}%
  \@dottedtocline{1}{0em}{2.2em}{\bfseries\hyperref[#1]{#2\quad #3}}{\hyperref[#1]{\pageref*{#1}}}%
}
\newcommand{\appcontentssubsection}[3]{%
  \@dottedtocline{2}{1.7em}{3.0em}{\hyperref[#1]{#2\quad #3}}{\hyperref[#1]{\pageref*{#1}}}%
}
\newcommand{\appcontentsunnumbered}[2]{%
  \@dottedtocline{2}{1.7em}{3.0em}{\hyperref[#1]{#2}}{\hyperref[#1]{\pageref*{#1}}}%
}
\makeatother

\phantomsection
\label{app:appendix_contents}
\begin{center}
  {\Large\bfseries Appendix for \avrb{}}\par
  \vspace{0.35ex}
  {\small \textit{When MLLMs Get Lost Between Rules and Instances in Abstract Visual Reasoning}}
\end{center}

\vspace{2.2ex}
\begingroup
\small
\renewcommand{\baselinestretch}{1.34}\selectfont
\setlength{\parskip}{0pt}
\setlength{\parindent}{0pt}
\appcontentssection{app:operation_primer}{A}{Benchmark Specification}
\appcontentssubsection{app:annotation_manual}{A.1}{Operation Definitions and Boundary Rules}
\appcontentssubsection{app:prf_spec}{A.2}{Shared-Stem P/R/F Specification}
\appcontentssubsection{app:s_schema}{A.3}{S1--S4 Annotation Schema and L3 Note}

\appcontentssection{app:dataset_qc}{B}{Dataset Construction and Statistics}
\appcontentssubsection{app:curation_pipeline}{B.1}{Curation and Construction Pipeline}
\appcontentssubsection{app:annotation_details}{B.2}{Annotation Protocol and Adjudication}
\appcontentssubsection{app:iaa}{B.3}{Annotation Audits and Golden-Set Monitoring}
\appcontentssubsection{app:datasheet}{B.4}{Dataset Card, Splits, and Statistics}

\appcontentssection{app:evaluation_protocols}{C}{Evaluation Protocol and Judge Calibration}
\appcontentssubsection{app:prompts}{C.1}{Direct and Thinking Prompt Library}
\appcontentssubsection{app:parsing}{C.2}{Output Parsing and Answer Extraction}
\appcontentssubsection{app:judge_calibration}{C.3}{L2 Judge Prompt and Human Calibration}
\appcontentssubsection{app:second_judge}{C.4}{Second-Judge Robustness on S3}
\appcontentssubsection{app:ssa_details}{C.5}{SSA Schema and Leakage Guards}

\appcontentssection{app:more_analysis}{D}{Additional Result Analyses}
\appcontentssubsection{app:full_leaderboard}{D.1}{Model Coverage and Per-Operation P/R/F}
\appcontentssubsection{app:binding_gap_ci}{D.2}{Strict Binding Gap Denominators and Confidence Intervals}
\appcontentssubsection{app:qwen_l2}{D.3}{Qwen3.5 L2 Stage and SSA Raw Values}
\appcontentssubsection{app:thinkgain_scatter}{D.4}{Paired Direct vs. Thinking Raw Values and Error Audit}
\appcontentssubsection{app:aux_metrics}{D.5}{Auxiliary Audits and Sanity Checks}

\appcontentssection{app:case}{E}{Qualitative Case Study Gallery}
\appcontentsunnumbered{app:case_index}{Case Index}
\appcontentssubsection{app:case_solved}{E.1}{Solved Reference Case}
\appcontentssubsection{app:case_cant_see}{E.2}{Case-Error: Can't-See}
\appcontentssubsection{app:case_cant_reason}{E.3}{Case-Error: Can't-Reason}
\appcontentssubsection{app:case_cant_bind}{E.4}{Case-Error: Can't-Bind}
\appcontentssubsection{app:case_worked_trace}{E.5}{Hard S1--S4 Worked Trace}
\appcontentssubsection{app:case_thinking}{E.6}{Case-Error: Direct vs. Thinking}

\appcontentssection{app:related_extended}{F}{Extended Related Work}

\appcontentssection{app:limitations_impact}{G}{Limitations and Broader Impact}
\appcontentssubsection{app:limitations}{G.1}{Limitations}
\appcontentssubsection{app:broader_impact}{G.2}{Broader Impact}
\endgroup

\clearpage

\section{Benchmark Specification}
\label{app:operation_primer}

\subsection{Operation Definitions and Boundary Rules}
\label{app:annotation_manual}
\label{app:chc_map}
The nine operation labels specify the visual operation a solver must infer, not a psychometric diagnosis of human ability. RI labels require abstract rule induction over observed panels; VP labels require mental transformation of a visual object or incomplete form. When an item admits multiple readings, annotators select the minimal operation needed to justify the answer and flag the item for adjudication.

\begin{table}[ht]
  \centering
  \caption{Operational definitions of the 9 AVR operation types. Family and
  type abbreviations are expanded in the table for readability. The
  rightmost column lists a reference CHC narrow ability purely for
  theoretical traceability; it is not the basis of label legitimacy.}
  \label{tab:taxonomy}
  \small
  \setlength{\tabcolsep}{3pt}
  \renewcommand{\arraystretch}{1.15}
  \begin{tabular}{@{}>{\raggedright\arraybackslash}p{0.18\textwidth}>{\raggedright\arraybackslash}p{0.23\textwidth}>{\raggedright\arraybackslash}p{0.40\textwidth}l@{}}
    \toprule
    Family (full name) & Type (full name) & Core operation & CHC ref\textsuperscript{$\dagger$} \\
    \midrule
    \multirow{5}{=}{\textbf{Rule Induction (RI)}}
      & RI-Pos (Position)    & Infer positional rule over a matrix axis                     & I \\
      & RI-Sty (Style)       & Infer stylistic rule (line / fill / shading)                 & I \\
      & RI-Attr (Attribute)  & Infer per-element attribute rule                             & I \\
      & RI-Qty (Quantity)    & Infer rule over element count / cardinality                  & I \\
      & RI-Rel (Relation)    & Infer relational rule between element pairs                  & I \\
    \midrule
    \multirow{4}{=}{\textbf{Visual Processing (VP)}}
      & VP-Fold (Folding)      & Match 2D net to its 3D folded form                          & Vz \\
      & VP-View (Viewpoint)    & Match object across viewpoint rotation                      & SR \\
      & VP-Rot (Rotation)      & Identify mental-rotation equivalence                        & SR \\
      & VP-Closure (Closure)   & Complete partially occluded / missing form                  & CS \\
    \bottomrule
  \end{tabular}
  \vspace{2pt}

  \footnotesize
  \textsuperscript{$\dagger$} CHC narrow reference included for theoretical
  traceability only; label validity comes from operation-definition
  reproducibility plus $\kappa$ audit (§2.2.1), not from CHC membership.
\end{table}

\begin{table}[htbp]
  \centering
  \footnotesize
  \setlength{\tabcolsep}{3pt}
  \renewcommand{\arraystretch}{1.08}
  \caption{Compact boundary rules for adjacent operation labels. Operation
  abbreviations are expanded in the first column.}
  \label{tab:operation_boundaries}
  \begin{tabular}{@{}>{\raggedright\arraybackslash}p{0.19\textwidth}>{\raggedright\arraybackslash}p{0.27\textwidth}>{\raggedright\arraybackslash}p{0.19\textwidth}>{\raggedright\arraybackslash}p{0.27\textwidth}@{}}
    \toprule
    Operation (full name) & Required evidence & Common neighbor & Boundary rule \\
    \midrule
    RI-Pos (Rule Induction Position) & Position changes over rows, columns, or slots & RI-Rel & Use RI-Pos when absolute or relative location is the primary state. \\
    RI-Sty (Rule Induction Style) & Line, fill, texture, or rendering style changes & RI-Attr & Use RI-Sty when appearance changes without changing object identity. \\
    RI-Attr (Rule Induction Attribute) & Object-level attributes such as color, shape, or size & RI-Sty / RI-Qty & Use RI-Attr when the attribute is bound to a specific element. \\
    RI-Qty (Rule Induction Quantity) & Count or cardinality progression & RI-Attr & Use RI-Qty when the solver must track how many elements exist. \\
    RI-Rel (Rule Induction Relation) & Pairwise or higher-order relations & RI-Pos & Use RI-Rel when the answer depends on correspondence, not only location. \\
    VP-Fold (Visual Processing Folding) & 2D net to 3D folded object matching & VP-View & Use VP-Fold when faces must be assembled from a net. \\
    VP-View (Visual Processing Viewpoint) & Object identity across viewpoint change & VP-Rot & Use VP-View when hidden/visible surfaces must be reconciled. \\
    VP-Rot (Visual Processing Rotation) & Rotation-equivalence or non-equivalence & VP-View & Use VP-Rot when the object is rigidly rotated. \\
    VP-Closure (Visual Processing Closure) & Missing or occluded contour completion & RI-Sty & Use VP-Closure when the missing form must be perceptually completed. \\
    \bottomrule
  \end{tabular}
\end{table}

\subsection{Shared-Stem P/R/F Specification}
\label{app:prf_spec}
Each visual stem yields three diagnostic probes over the same evidence. P probes check whether the model identifies local visual entities or attributes; R probes check whether it selects the abstract rule; F tasks require binding that rule to the target answer instance. This shared-stem structure is the denominator for the R--F chasm, the stem-level failure decomposition, and the strict Binding Gap.

\begin{table}[htbp]
  \centering
  \small
  \setlength{\tabcolsep}{4pt}
  \renewcommand{\arraystretch}{1.1}
  \caption{Schema fields for a shared-stem diagnostic unit.}
  \label{tab:prf_schema}
  \begin{tabular}{@{}p{0.14\textwidth}p{0.30\textwidth}p{0.46\textwidth}@{}}
    \toprule
    Field & Meaning & Example content \\
    \midrule
    \texttt{stem\_id} & Shared visual stem key & One curated AVR stem reused by P, R, and F probes. \\
    \texttt{operation} & Primary operation label & One of the nine RI/VP operations. \\
    \texttt{P\_probe} & Perception question & Identify visible entities, positions, attributes, or counts. \\
    \texttt{R\_probe} & Rule question & Select or describe the transformation governing the stem. \\
    \texttt{F\_task} & Full AVR item & Choose the answer option that satisfies the inferred rule. \\
    \texttt{S1--S4} & Process targets & Encode, infer, map, and apply targets used by L2 judging and \ssa{}. \\
    \bottomrule
  \end{tabular}
\end{table}

\begin{figure}[htbp]
  \centering
  \includegraphics[width=0.94\textwidth]{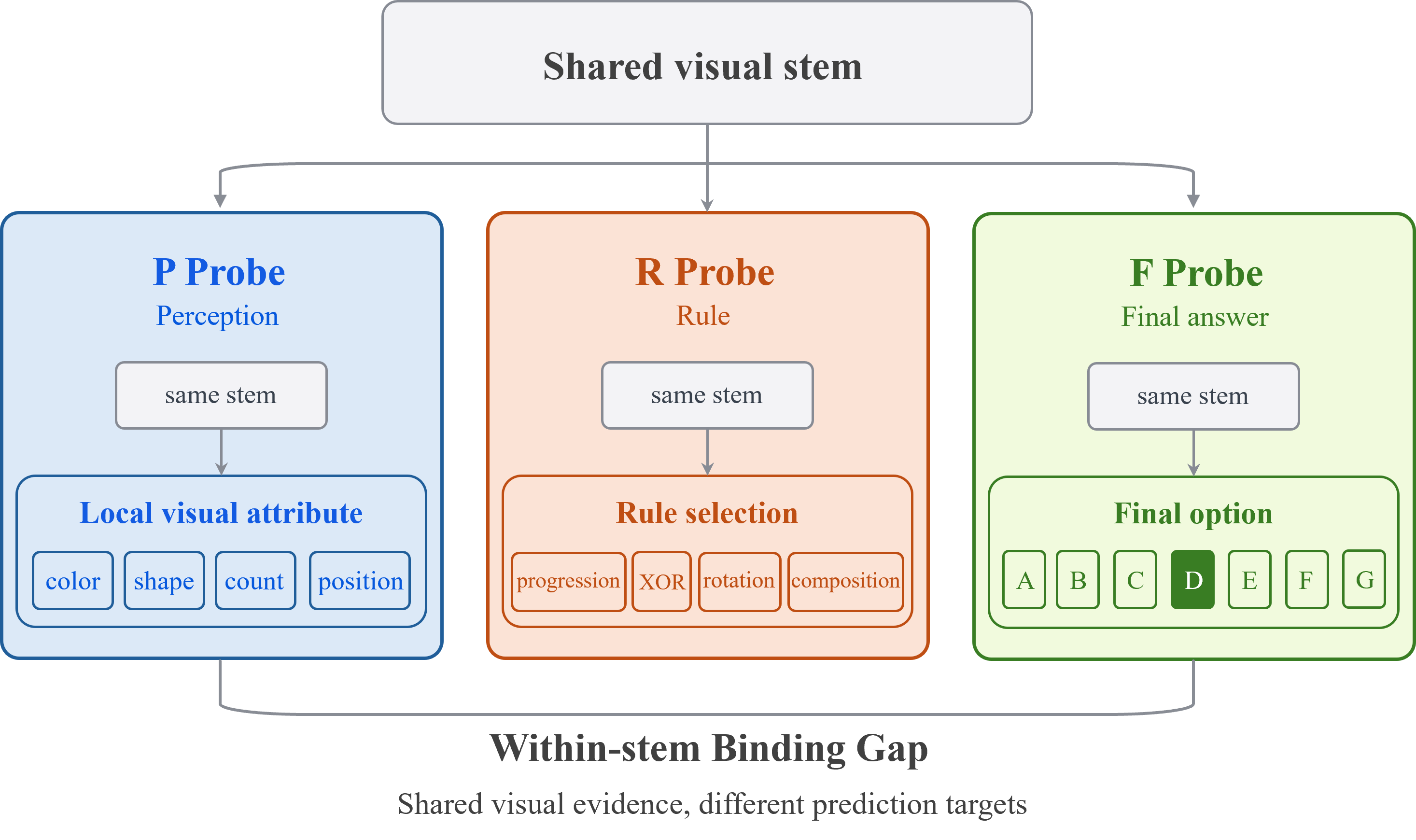}
  \caption{Shared-stem P/R/F diagnostic card. P probes, R probes, and F tasks are derived from the same visual stem, so they share visual evidence while requiring different outputs: local visual attributes, rule selection, and the final answer option. This shared evidence makes the within-stem Binding Gap well-defined.}
  \label{fig:app_prf_shared_stem_card}
\end{figure}

\subsection{S1--S4 Annotation Schema and L3 Note}
\label{app:s_schema}
\label{app:l3_tags}
The S1--S4 schema defines the expected solution path for an F item. L3 AttrTag is obtained by crossing the first failed S1--S4 stage with the perception-load tag. It is retained as behavioral metadata for future analysis; the current paper does not report separate L3-level results, and L3 is not used as an independent evidence source for the main findings.

\begin{table}[htbp]
  \centering
  \small
  \setlength{\tabcolsep}{5pt}
  \renewcommand{\arraystretch}{1.1}
  \caption{S1--S4 annotation boundaries.}
  \label{tab:s1s4_schema}
  \begin{tabular}{@{}p{0.13\textwidth}p{0.22\textwidth}p{0.30\textwidth}p{0.25\textwidth}@{}}
    \toprule
    Stage & Target & Correct evidence & Typical failure \\
    \midrule
    S1 Encode & Visual entities and attributes & Objects, positions, colors, counts, contours & Missing or hallucinated visual element \\
    S2 Infer & Abstract rule & Transformation, relation, or progression & Wrong rule family or missing rule \\
    S3 Map & Rule-to-instance alignment & Correct target slot, correspondence, or option mapping & Correct rule, wrong answer slot \\
    S4 Apply & Final answer & Correct option after applying the mapping & Calculation or option-selection error \\
    \bottomrule
  \end{tabular}
\end{table}

\section{Dataset Construction and Statistics}
\label{app:dataset_qc}

\subsection{Curation and Construction Pipeline}
\label{app:curation_pipeline}
\avrb{} does not release copied public puzzle images. Candidate stems are retained only after provenance screening, near-duplicate filtering, and human review, while preserving the abstract operation, relation structure, and answer logic required by the benchmark taxonomy. Surface style, layout, and option identity are checked before P/R/F proposal, human review, adjudication, filtering, and split assignment.

\begin{table}[htbp]
  \centering
  \small
  \setlength{\tabcolsep}{5pt}
  \renewcommand{\arraystretch}{1.08}
  \caption{Construction pipeline and audit question per stage.}
  \label{tab:construction_pipeline}
  \begin{tabular}{@{}p{0.20\textwidth}p{0.67\textwidth}@{}}
    \toprule
    Stage & Audit question \\
    \midrule
    Source selection & Does the source provide a knowledge-light AVR structure? \\
    Curation & Does the released stem pass provenance screening while preserving abstract answer logic? \\
    P/R/F proposal & Are perception, rule, and full probes tied to the same visual evidence? \\
    Human review & Do annotators agree that probes are valid and answerable? \\
    Adjudication & Are operation, difficulty, S1--S4 targets, and options resolved by an expert? \\
    Filtering & Are duplicate, malformed, or shortcut-prone items removed? \\
    Split assignment & Are public and hidden splits separated with answer randomization? \\
    \bottomrule
  \end{tabular}
\end{table}

\paragraph{Stem provenance.} Table~\ref{tab:provenance} lists the source of each stem. 1{,}011 stems (44.0\%) are newly constructed by our annotators, and 1{,}287 (56.0\%) are adapted from four license-screened public AVR datasets. Adapted stems keep only the abstract rule and answer logic. Their surface, layout, symbols, and options are rebuilt, and their operation labels, P/R/F probes, and S1--S4 targets come from our annotation pipeline. Released stems pass pHash and CLIP near-duplicate screening against all four source datasets.

\begin{table}[htbp]
  \centering
  \small
  \setlength{\tabcolsep}{6pt}
  \renewcommand{\arraystretch}{1.08}
  \caption{Stem provenance. Adapted stems keep only the abstract rule and answer logic; surface, layout, symbols, and options are rebuilt.}
  \label{tab:provenance}
  \begin{tabular}{@{}lrr@{}}
    \toprule
    Source & Stems & Share \\
    \midrule
    Newly constructed & 1{,}011 & 44.0\% \\
    Adapted from public AVR datasets & 1{,}287 & 56.0\% \\
    \midrule
    Total & 2{,}298 & 100.0\% \\
    \bottomrule
  \end{tabular}
\end{table}

\subsection{Annotation Protocol and Adjudication}
\label{app:annotation_details}
Annotators are trained on the operation manual, shared-stem P/R/F validity rules, and S1--S4 boundary cases before labeling production items. GPT-5 and Claude proposals are used only as drafts for candidate probes and process targets; final authority remains with two human annotators per item plus an expert adjudicator. Annotation spanned roughly 1.5 months for 2--3 trained annotators and one adjudicator.

\begin{table}[htbp]
  \centering
  \small
  \setlength{\tabcolsep}{5pt}
  \renewcommand{\arraystretch}{1.08}
  \caption{Adjudication rules applied during annotation.}
  \label{tab:adjudication_rules}
  \begin{tabular}{@{}p{0.22\textwidth}p{0.31\textwidth}p{0.34\textwidth}@{}}
    \toprule
    Disagreement type & Resolution rule & Failure action \\
    \midrule
    Operation label & Prefer the minimal operation needed to solve F & Escalate ambiguous multi-operation items \\
    P/R/F validity & Require answerability from the shared visual stem & Rewrite or remove malformed probes \\
    S1--S4 target & Match the first necessary step in the intended solution & Rewrite target text and re-check answer key \\
    Option identity & Verify answer after option shuffling & Remove shortcut-prone options \\
    \bottomrule
  \end{tabular}
\end{table}

\subsection{Annotation Audits and Golden-Set Monitoring}
\label{app:iaa}
Annotation quality is monitored at the dimension level rather than only at final answer accuracy. The release package includes the annotator monitoring logs; Table~\ref{tab:iaa_template} reports the frozen deterministic audit checks used by this submission.

\begin{table}[htbp]
  \centering
  \small
  \setlength{\tabcolsep}{5pt}
  \renewcommand{\arraystretch}{1.08}
  \caption{Frozen annotation and dataset audit checks.}
  \label{tab:iaa_template}
  \begin{tabular}{@{}p{0.25\textwidth}p{0.22\textwidth}p{0.15\textwidth}p{0.28\textwidth}@{}}
    \toprule
    Dimension & Audit metric & Value & Main issue checked \\
    \midrule
    Operation label & Confidence audit & min 0.71; mean 0.923 & Boundary operation \\
    P/R/F validity & Schema completeness & 2{,}298/2{,}298 stems & Malformed or untied probes \\
    S1--S4 target & Target completeness & 2{,}298/2{,}298 F items & Missing stage target \\
    Answer key & Letter-balance audit & max share 26.3\% & Option-level shortcut \\
    \bottomrule
  \end{tabular}
\end{table}

\subsection{Dataset Card, Splits, and Statistics}
\label{app:datasheet}
\label{app:data_stats}
The released result-bearing split contains 2{,}298 stems and 19{,}533 tasks: 14{,}937 P probes, 2{,}298 R probes, and 2{,}298 F items. Intended use is diagnostic evaluation of MLLM abstract visual reasoning; the dataset should not be treated as a psychometric test of humans nor as a training corpus for memorizing item templates.

\begin{table}[htbp]
  \centering
  \small
  \setlength{\tabcolsep}{5pt}
  \renewcommand{\arraystretch}{1.08}
  \caption{Dataset card summary.}
  \label{tab:dataset_card_appendix}
  \begin{tabular}{@{}p{0.24\textwidth}p{0.62\textwidth}@{}}
    \toprule
    Field & Summary \\
    \midrule
    Released result-bearing split & 2{,}298 stems; 19{,}533 tasks; 14{,}937 P / 2{,}298 R / 2{,}298 F. \\
    Release format & One Hugging Face record per stem, bundling the stem image, F task, R probe, P probes, and S1--S4 annotations. \\
    Image source policy & 1{,}011 newly constructed stems and 1{,}287 stems adapted from four license-screened public AVR datasets with rebuilt surface, layout, symbols, and options (Table~\ref{tab:provenance}); all stems are provenance-screened with near-duplicate filtering. \\
    Intended use & Diagnostic benchmarking of MLLM abstract visual reasoning. \\
    Known limitations & English prompts, finite operation taxonomy, and behavioral rather than mechanistic stage labels. \\
    \bottomrule
  \end{tabular}
\end{table}

\begin{figure}[H]
  \centering
  \includegraphics[width=0.86\textwidth]{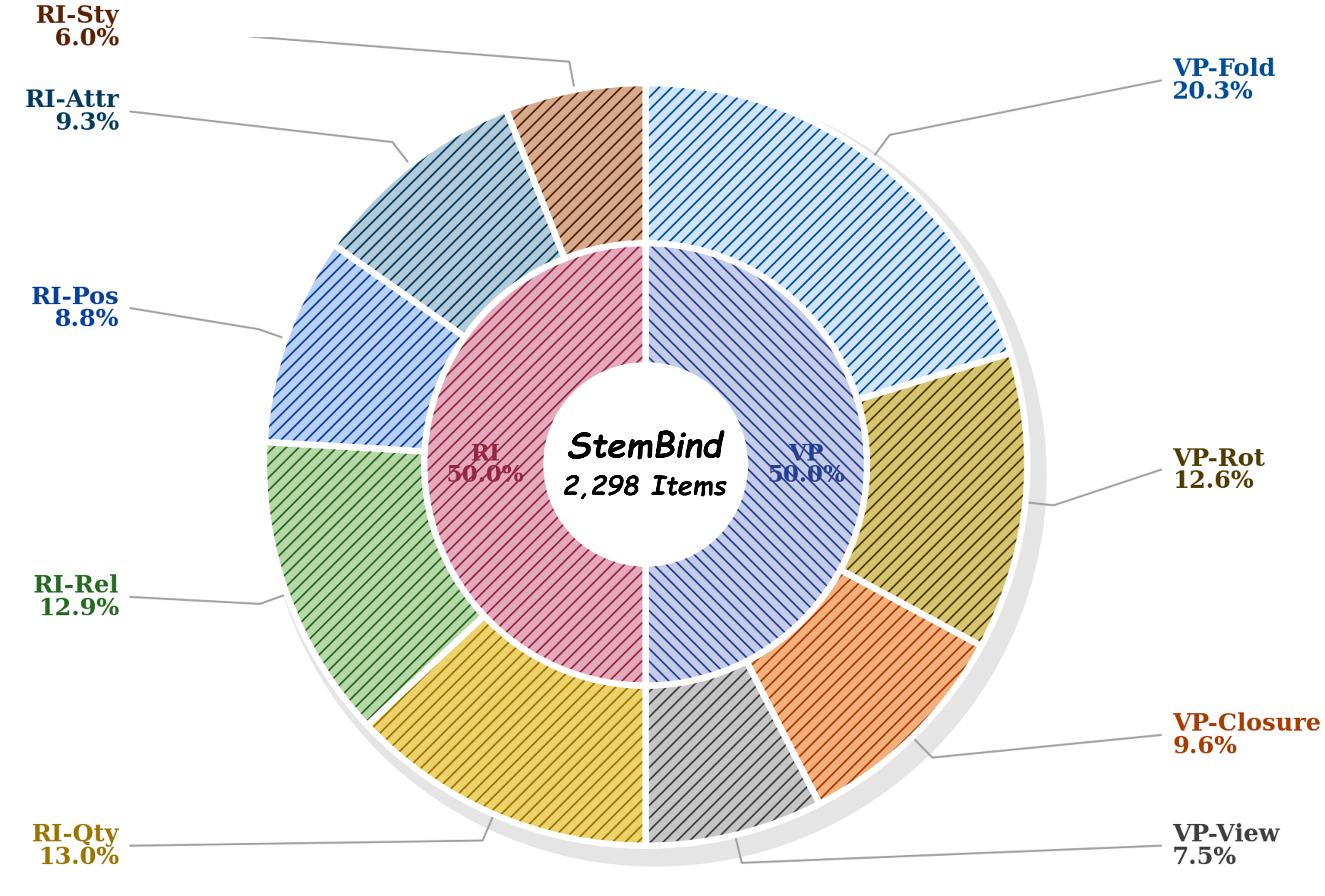}
  \caption{Operation distribution across the 2{,}298 result-bearing stems. The inner ring aggregates RI/VP families; the outer ring shows the nine operation labels used throughout \avrb{}.}
  \label{fig:app_dataset_operation_distribution}
\end{figure}

\begin{figure}[H]
  \centering
  \includegraphics[width=0.86\textwidth]{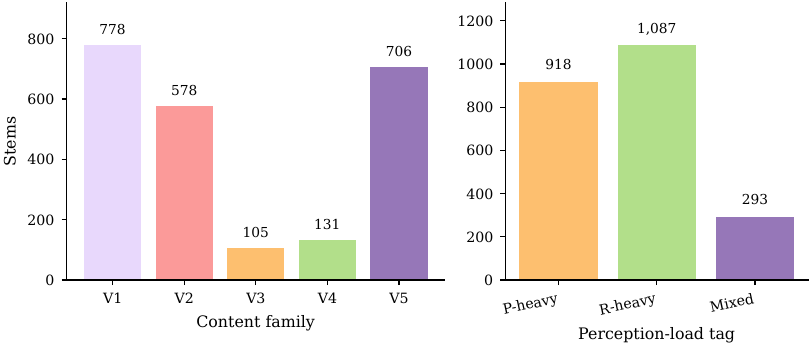}
  \caption{Visual-content and perception-load distributions. Probe level (P/R/F) is distinct from the item-level perception-load tag used for L3 metadata.}
  \label{fig:app_dataset_content_perception_distribution}
\end{figure}

\begin{figure}[H]
  \centering
  \includegraphics[width=0.86\textwidth]{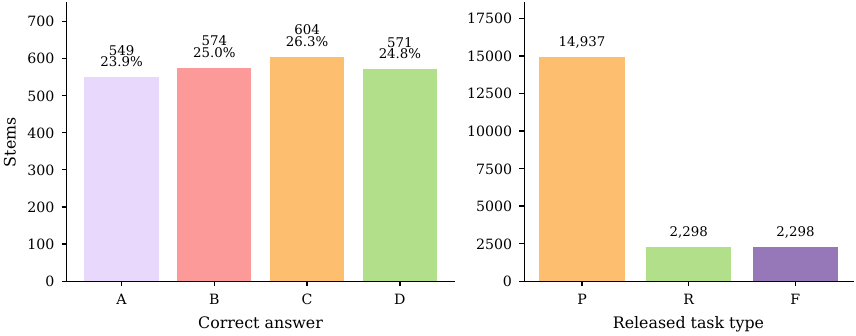}
  \caption{Answer-option balance and released P/R/F task counts. This audit checks that answer letters are approximately balanced and that the public release exposes 14{,}937 P probes, 2{,}298 R probes, and 2{,}298 F items.}
  \label{fig:app_dataset_answer_probe_distribution}
\end{figure}

\begin{figure}[H]
  \centering
  \includegraphics[width=0.90\textwidth]{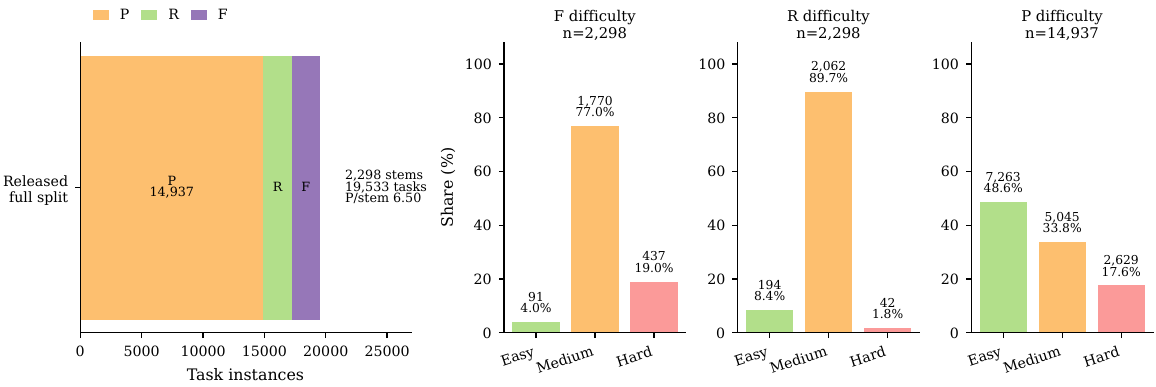}
  \caption{Released full-split statistics and easy/medium/hard distributions for F, R, and P across the 2{,}298 stems and 19{,}533 P/R/F tasks.}
  \label{fig:app_dataset_english_annotation_difficulty}
\end{figure}

\section{Evaluation Protocol and Judge Calibration}
\label{app:evaluation_protocols}

\subsection{Direct and Thinking Prompt Library}
\label{app:prompts}
\label{app:compute}
All benchmark rows use English stems, full-image input, temperature 0, and fixed max-token budgets. Direct prompts ask the model to solve the item and place the final choice in \texttt{<ANSWER>X</ANSWER>}. Thinking prompts use the same image and answer space but allow longer reasoning~\citep{wei2022cot,zhou2023least,wang2023plan}; thinking rows are paired diagnostics and are not mixed into direct-mode leaderboards or family scaling.

\begin{table}[htbp]
  \centering
  \small
  \setlength{\tabcolsep}{5pt}
  \renewcommand{\arraystretch}{1.08}
  \caption{Prompt variants used by the evaluation. Verbatim prompt text is released with the code package.}
  \label{tab:prompt_variants}
  \begin{tabular}{@{}p{0.18\textwidth}p{0.34\textwidth}p{0.34\textwidth}@{}}
    \toprule
    Variant & Instruction shape & Use in paper \\
    \midrule
    Direct P/R/F & Answer the probe from the image and place the final choice in the answer tag & Main leaderboard and P/R/F diagnostics \\
    Thinking P/R/F & Reason more explicitly before the same final answer tag & Paired \tg{} diagnostics only \\
    L2 trace & Free reasoning trace plus final answer tag & Input to S1--S4 judge \\
    \ssa{} H0--H4 & Same F task with cumulative verified stage prefill & Stage intervention diagnostics \\
    \bottomrule
  \end{tabular}
\end{table}

Open-source and local model runs were executed on a server with 8 NVIDIA H100 GPUs (80\,GB memory each). Proprietary model rows were evaluated through provider APIs; provider-side hardware is not observable.

\subsection{Output Parsing and Answer Extraction}
\label{app:parsing}
The parser first searches for a \texttt{<ANSWER>X</ANSWER>} tag. If the tag is absent, it falls back only when a unique answer letter can be unambiguously normalized from the final response. Multiple answer letters, missing final choices, and malformed outputs are marked invalid and counted wrong.

\begin{table}[htbp]
  \centering
  \small
  \setlength{\tabcolsep}{5pt}
  \renewcommand{\arraystretch}{1.08}
  \caption{Answer extraction rules.}
  \label{tab:parser_rules}
  \begin{tabular}{@{}p{0.26\textwidth}p{0.28\textwidth}p{0.32\textwidth}@{}}
    \toprule
    Output pattern & Parser action & Status \\
    \midrule
    Tagged single answer & Accept the normalized letter & Valid \\
    Untagged unique final answer & Normalize with fallback & Valid after fallback \\
    Multiple answer letters & Mark invalid & Wrong \\
    No final answer & Mark invalid & Wrong \\
    \bottomrule
  \end{tabular}
\end{table}

\subsection{L2 Judge Prompt and Human Calibration}
\label{app:judge_calibration}
The L2 judge takes a model trace, the final answer, the S1--S4 ground truth, and task metadata, then emits binary correctness for each stage and the first-failure stage. The judge is GPT-4o at temperature 0; it is not a member of the evaluated Qwen3.5 family and is not the Qwen2.5-7B used in earlier StepAcc pipelines. The decision rule walks S1$\rightarrow$S4 in order and stops at the first stage marked incorrect.

\paragraph{Calibration sample.} We draw a fixed stratified sample of $n{=}180$ Qwen3.5-9B F traces, with 20 traces per operation and coverage over three model-outcome bins (model F correct, model F wrong with R correct, model F wrong with R wrong). Three trained annotators (A1, A2, A3) independently assign S1/S2/S3/S4 binary correctness on the same traces under one per-stage protocol. All three are blind to the judge labels. A2 and A3 are also blind to A1 and to each other. Cohen's $\kappa$ is computed per stage for each judge--annotator pair and for the A2--A3 pair, which serves as a human--human ceiling. The sampled item list is frozen in \texttt{paper/calibration/calibration\_sample\_180.csv}.

\begin{figure}[htbp]
  \centering
  \includegraphics[width=0.62\linewidth]{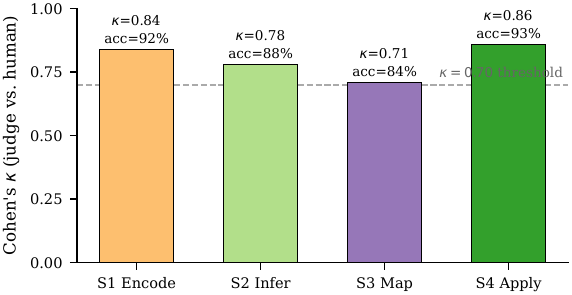}
  \caption{GPT-4o L2 judge vs.\ annotator A1, Cohen's $\kappa$ across S1--S4 on the 180-item calibration set. All four stages clear the $\kappa{=}0.70$ reliability threshold; S3 Map is the lowest, consistent with the boundary cases noted in Sec.~\ref{app:s_schema}. Table~\ref{tab:judge_calibration} adds A2, A3, and the human--human ceiling.}
  \label{fig:judge_calibration}
\end{figure}

\begin{table}[htbp]
  \centering
  \small
  \setlength{\tabcolsep}{6pt}
  \renewcommand{\arraystretch}{1.08}
  \caption{Judge and human calibration on 180 stratified F traces. Rows report pairwise Cohen's $\kappa$ per stage; the last row reports raw agreement between the judge and A1.}
  \label{tab:judge_calibration}
  \begin{tabular}{@{}lcccc@{}}
    \toprule
    Pairing & S1 Encode & S2 Infer & S3 Map & S4 Apply \\
    \midrule
    Judge vs.\ A1 & 0.84 & 0.78 & 0.71 & 0.86 \\
    Judge vs.\ A2 & 0.83 & 0.77 & 0.72 & 0.85 \\
    Judge vs.\ A3 & 0.81 & 0.75 & 0.69 & 0.83 \\
    A2 vs.\ A3 (human ceiling) & 0.89 & 0.83 & 0.80 & 0.92 \\
    \midrule
    Accuracy agreement (judge vs.\ A1) & 92\% & 88\% & 84\% & 93\% \\
    \bottomrule
  \end{tabular}
\end{table}

\paragraph{Reading the table.} S1 and S4 are highest because the underlying judgments are mostly extractive (visible entities or option letters). S3 is lowest in all four pairings, including between A2 and A3, because the rule-to-instance mapping admits multiple equivalent verbal phrasings. The judge's gap to the human ceiling at S3 exceeds its gap at the other stages by at most $0.03$ in $\kappa$, so the lower S3 agreement comes mainly from the stage itself rather than from the judge. The aggregate claim in the main text (S3 is systematically weakest and the largest \ssa{} gain appears at H2$\rightarrow$H3) does not depend on resolving every individual S3 disagreement.

\paragraph{First-failure attribution under human labels.} On the 115 F-wrong traces of the sample, the A2/A3 consensus labels make S3 the leading first-failure stage at $45\%$, close to the $48\%$ under the judge labels on the same traces.

\subsection{Second-Judge Robustness on S3}
\label{app:second_judge}
S3 carries the lowest GPT-4o vs.\ human $\kappa$ in Tab.~\ref{tab:judge_calibration} and also anchors the behavioral binding claim in the main text. To check that the S3 ranking is not an artifact of a single judge, we re-score the same 180-trace calibration sample and the full Qwen3.5 F split with a second judge, Qwen3.5-Plus, run at temperature 0 with the same S1--S4 prompt, the same first-failure decision rule, and no access to the GPT-4o stage labels. Qwen3.5-Plus is a closed model from a different release line than the open-weights Qwen3.5 family on the leaderboard, so it does not score traces produced by itself. We treat Qwen3.5-Plus as a secondary verifier rather than a primary judge; the calibration target is convergence on the S3 ranking and the family-level claim, not exact match on every per-item label.

\begin{table}[htbp]
  \centering
  \small
  \setlength{\tabcolsep}{6pt}
  \renewcommand{\arraystretch}{1.08}
  \caption{Second-judge robustness on S3. The first row reuses Tab.~\ref{tab:judge_calibration}. Inter-judge agreement is computed on the same 180-trace calibration set; ranking stability uses the seven Qwen3.5 family rows from Tab.~\ref{tab:qwen_l2_template}.}
  \label{tab:second_judge}
  \begin{tabular}{@{}lcc@{}}
    \toprule
    Metric & GPT-4o & Qwen3.5-Plus \\
    \midrule
    Cohen's $\kappa$ vs.\ A1 (S3) & 0.71 & 0.69 \\
    Inter-judge $\kappa$ on S3 ($n{=}180$) & \multicolumn{2}{c}{0.74} \\
    Qwen3.5 family mean S3 StepAcc & 30.5\% & 31.8\% \\
    S3-weakest-stage rows preserved & 7/7 & 6/7 \\
    Spearman $\rho$ on S3 ranking (7 rows) & \multicolumn{2}{c}{0.93} \\
    \bottomrule
  \end{tabular}
\end{table}

The S3 StepAcc family mean shifts by 1.3 percentage points under the judge swap, and the Spearman $\rho$ of 0.93 across the seven family rows indicates that S3-based model ordering is preserved up to two small adjacent moves. The S3-weakest-stage call is preserved on 6 of 7 rows; the single swap occurs on Qwen3.5-0.8B, where all four stage StepAcc values cluster near floor (S1=24.1\%, S2=22.8\%, S3=16.0\%, S4=35.4\%) and a few S2/S3 boundary calls move S2 below S3 under the second judge. Per-item disagreements remain concentrated on S3 boundary cases, consistent with App.~\ref{app:judge_calibration}, but the aggregate behavioral binding claim is not a property of the GPT-4o judge alone.

\subsection{SSA Schema and Leakage Guards}
\label{app:ssa_details}
\ssa{} injects verified stage information cumulatively from H0 to H4. No condition includes the final answer letter, and each level forbids downstream content beyond the intended stage. The irrelevant-S1 control checks whether improvements come from generic extra text rather than stage-specific content.

\begin{table}[t]
  \centering
  \small
  \setlength{\tabcolsep}{4pt}
  \renewcommand{\arraystretch}{1.15}
  \caption{\ssa{} prefill schema and leakage guards.  Conditions are
  cumulative: each level adds one verified S-stage while forbidding
  downstream information and the final answer.}
  \label{tab:ssa_schema}
  \begin{tabular}{@{}lp{0.42\textwidth}p{0.36\textwidth}@{}}
    \toprule
    Cond. & Injected prefill content
          & Must NOT appear (leakage guard) \\
    \midrule
    \textbf{H0} Baseline
      & None; raw stem only.
      & Any S1--S4 intermediate; final answer. \\
    \textbf{H1} +S1
      & Verified S1 \emph{scene graph}: element list, attributes, grid
        coordinates.
      & S2 rule hypotheses; S3 alignment; S4 target; final answer. \\
    \textbf{H2} +S1,S2
      & H1 content plus verified S2 \emph{inferred rules} (natural-language
        rule statements).
      & S3 mapping; S4 target instance; final answer. \\
    \textbf{H3} +S1,S2,S3
      & H2 content plus verified S3 \emph{alignment} between the inferred
        rule and candidate slot(s).
      & S4 target instance; final answer. \\
    \textbf{H4} +S1--S4
      & H3 content plus verified S4 \emph{target instance description}
        (element-level, \emph{not} the option letter).
      & The final \texttt{<ANSWER>} letter. \\
    \bottomrule
  \end{tabular}
\end{table}

\section{Additional Result Analyses}
\label{app:more_analysis}

\subsection{Model Coverage and Per-Operation P/R/F}
\label{app:full_leaderboard}
\label{app:per_operation_results}
The main text reports the direct-mode leaderboard. Table~\ref{tab:model_tiers} fixes which rows feed the leaderboard, family scaling, \tg{} pairs, and \ssa{}. Per-operation results test whether the R--F chasm is a single-operation artifact; the paper reports operation-level summaries by model group, while the complete 24$\times$9 matrix is released as CSV to avoid spending appendix pages on a dense table.

\begin{table}[t]
  \centering
  \footnotesize
  \setlength{\tabcolsep}{3.5pt}
  \renewcommand{\arraystretch}{1.08}
  \caption{Model coverage by experiment.  The main leaderboard uses
  five closed/hosted frontier models, two open-source standalone frontier
  models, and three open-source scaling families.
  Family-scaling curves use the full Qwen3.5, InternVL3.5, and Gemma~4 size
  ranges; \tg{} is restricted to rows with a matched direct / thinking
  control in Table~\ref{tab:main_leaderboard}.}
  \label{tab:model_tiers}
  \begin{tabular}{@{}>{\raggedright\arraybackslash}p{0.16\textwidth}
                  >{\raggedright\arraybackslash}p{0.19\textwidth}
                  >{\raggedright\arraybackslash}p{0.58\textwidth}@{}}
    \toprule
    Block & Purpose & Models \\
    \midrule
    Main leaderboard
      & Closed / hosted frontier
      & GPT-5.4, Gemini-3.1-Pro, Claude-Opus-4.7, Qwen3.5-Plus,
        grok-4.2-beta \\
      & Open-source models
      & Kimi-K2.5; GLM-4.5V;
        Qwen3.5 \{0.8B, 2B, 4B, 9B, 27B, 35B-A3B, 122B-A10B\};
        InternVL3.5 \{1B, 2B, 4B, 8B, 14B, 30B-A3B\};
        Gemma~4 \{E2B-it, E4B-it, 26B-A4B-it, 31B-it\} \\
    \midrule
    Family scaling
      & Qwen3.5 family
      & 0.8B, 2B, 4B, 9B, 27B, 35B-A3B, 122B-A10B \\
      & InternVL3.5 family
      & 1B, 2B, 4B, 8B, 14B, 30B-A3B \\
      & Gemma~4 family
      & E2B-it, E4B-it, 26B-A4B-it, 31B-it \\
    \midrule
    \tg{} pairs
      & Matched direct/thinking controls
      & GPT-5.4, Gemini-3.1-Pro, Claude-Opus-4.7, Qwen3.5-Plus,
        grok-4.2-beta; Qwen3.5 \{0.8B, 2B, 4B, 9B, 27B\} \\
    \midrule
    SSA full split
      & Stage-level intervention
      & Qwen3.5 \{0.8B, 2B, 4B, 9B, 27B, 35B-A3B, 122B-A10B\} \\
    \bottomrule
  \end{tabular}
\end{table}

\begin{figure}[H]
  \centering
  \includegraphics[width=0.95\textwidth]{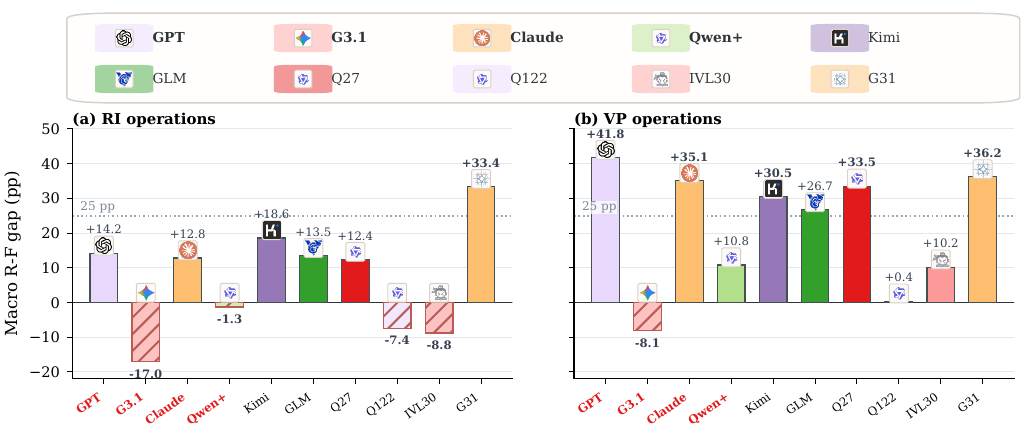}
  \caption{Operation-family R--F chasm. Bars summarize macro R--F gaps separately over RI and VP operations for representative rows, showing that the chasm is not driven by a single operation family.}
  \label{fig:app_op_family_chasm}
\end{figure}

\begin{table}[htbp]
  \centering
  \small
  \setlength{\tabcolsep}{5pt}
  \renewcommand{\arraystretch}{1.08}
  \caption{RI/VP family P/R/F group means (\%). The complete model-by-operation matrix is released as CSV.}
  \label{tab:per_operation_template}
  \begin{tabular}{@{}llcccc@{}}
    \toprule
    Model group & Operation block & P & R & F & R--F gap \\
    \midrule
    Proprietary frontier & RI & 60.0 & 35.4 & 34.9 & 0.5 \\
    Proprietary frontier & VP & 60.6 & 51.7 & 33.2 & 18.5 \\
    Qwen3.5 family & RI & 62.8 & 43.3 & 36.4 & 6.9 \\
    Qwen3.5 family & VP & 63.6 & 64.6 & 36.2 & 28.4 \\
    InternVL3.5 family & RI & 36.5 & 19.9 & 26.2 & -6.3 \\
    InternVL3.5 family & VP & 41.4 & 35.8 & 25.3 & 10.5 \\
    Gemma 4 family & RI & 56.2 & 56.2 & 31.2 & 24.9 \\
    Gemma 4 family & VP & 73.4 & 59.6 & 31.8 & 27.7 \\
    \bottomrule
  \end{tabular}
\end{table}

\subsection{Strict Binding Gap Denominators and Confidence Intervals}
\label{app:binding_gap_ci}
The strict Binding Gap conditions on stems where every P probe and the R probe are correct on the same stem. This denominator is intentionally stricter than the can't-bind category, because it asks whether F remains difficult even after full observed P and R success. Each stem carries $6.50$ annotated P probes on average, and all of them must be correct. The unit of analysis is a model--stem observation, so one stem contributes once per model that meets the condition.

\begin{table}[htbp]
  \centering
  \footnotesize
  \setlength{\tabcolsep}{3pt}
  \renewcommand{\arraystretch}{1.08}
  \caption{Strict Binding Gap denominators with 95\% bootstrap CIs (open-full split).}
  \label{tab:binding_gap_ci}
  \begin{tabular}{@{}p{0.34\textwidth}rrrr@{}}
    \toprule
    Model / group & \makecell{Strict $P{=}1,$\\$R{=}1$ stems} & \makecell{$P(F{=}1\mid$\\$P,R)$} & Binding Gap & 95\% CI \\
    \midrule
    Open-full Qwen3.5 / Gemma 4 pool & 1{,}385 & 48.8\% & 0.512 & [0.486, 0.538] \\
    Qwen3.5-27B & 261 & 53.3\% & 0.467 & [0.407, 0.528] \\
    Qwen3.5-9B & 191 & 51.3\% & 0.487 & [0.416, 0.560] \\
    Gemma-4-31B-it & 202 & 45.5\% & 0.545 & [0.475, 0.615] \\
    Gemma-4-26B-A4B-it & 182 & 46.2\% & 0.538 & [0.465, 0.609] \\
    \bottomrule
  \end{tabular}
\end{table}

\paragraph{Logistic check.} As a guard against reading the Binding Gap as a mechanistic ``synergy deficit'', we fit
\begin{equation}
  F \sim P + R + P{:}R + \alpha_m,
\end{equation}
with model fixed effects $\alpha_m$. The interaction is positive but small ($\hat{\gamma}{=}0.25$, $z{=}2.37$, $p{=}0.018$) and adds only $0.0002$ McFadden pseudo-$R^2$ over P/R main effects. Joint P/R correctness helps, but it does not eliminate the empirical conditional failure rate.

\paragraph{Derivation-checked R criterion.} The main accuracies score R by rule selection (L1). For a stricter test, a response counts as R-correct only if it selects the correct rule and states a rule that matches the gold S2 (Infer) annotation of the item. On the 1{,}100 pooled observations that meet this criterion and pass all P probes, F is still wrong $48\%$ of the time (95\% CI $[0.45, 0.51]$), close to the $0.512$ under L1 selection. Since the model must state the rule itself, these cases cannot come from eliminating options.

\subsection{Qwen3.5 L2 Stage and \ssa{} Raw Values}
\label{app:qwen_l2}
\label{app:qwen_ssa}
The Qwen3.5 L2 diagnostic covers the full F split. S3 Map is the weakest stage on every reported row. Mean StepAcc values (S1=43.5\%, S2=42.2\%, S3=30.5\%, S4=45.9\%) match the main-text summary.

\begin{table}[htbp]
  \centering
  \small
  \setlength{\tabcolsep}{5pt}
  \renewcommand{\arraystretch}{1.08}
  \caption{Qwen3.5 L2 stage StepAcc on the full F split.}
  \label{tab:qwen_l2_template}
  \begin{tabular}{@{}lccccc@{}}
    \toprule
    Model & S1 StepAcc & S2 StepAcc & S3 StepAcc & S4 StepAcc & Weakest stage \\
    \midrule
    Qwen3.5 family mean & 43.5\% & 42.2\% & 30.5\% & 45.9\% & S3 \\
    Qwen3.5-0.8B & 24.1\% & 22.8\% & 16.0\% & 35.4\% & S3 \\
    Qwen3.5-2B & 37.8\% & 35.8\% & 24.7\% & 42.0\% & S3 \\
    Qwen3.5-4B & 43.2\% & 42.4\% & 30.5\% & 46.3\% & S3 \\
    Qwen3.5-9B & 45.6\% & 44.6\% & 32.8\% & 48.2\% & S3 \\
    Qwen3.5-27B & 53.0\% & 51.6\% & 37.9\% & 51.0\% & S3 \\
    Qwen3.5-35B-A3B & 49.7\% & 48.5\% & 34.9\% & 48.4\% & S3 \\
    Qwen3.5-122B-A10B & 51.0\% & 50.0\% & 36.4\% & 49.7\% & S3 \\
    \bottomrule
  \end{tabular}
\end{table}

\begin{table}[htbp]
  \centering
  \small
  \setlength{\tabcolsep}{4pt}
  \renewcommand{\arraystretch}{1.08}
  \caption{Qwen3.5 \ssa{} raw values; H1 alone is negative on every row, the largest gain is at H2$\rightarrow$H3, and the irrelevant-S1 control stays below H0.}
  \label{tab:qwen_ssa_template}
  \begin{tabular}{@{}lccccccc@{}}
    \toprule
    Model & H0 & H1 & H2 & H3 & H4 & Largest gain & Irrelevant-S1 \\
    \midrule
    Qwen3.5 family mean & 36.0 & 30.0 & 38.1 & 49.0 & 53.4 & H2$\rightarrow$H3 & 26.7 \\
    Qwen3.5-0.8B & 30.2 & 22.1 & 29.0 & 36.7 & 40.3 & H2$\rightarrow$H3 & 22.8 \\
    Qwen3.5-2B & 35.9 & 26.4 & 32.9 & 43.3 & 46.2 & H2$\rightarrow$H3 & 22.9 \\
    Qwen3.5-4B & 33.8 & 31.5 & 39.3 & 49.9 & 55.8 & H2$\rightarrow$H3 & 26.0 \\
    Qwen3.5-9B & 37.4 & 32.4 & 40.8 & 52.7 & 56.2 & H2$\rightarrow$H3 & 29.0 \\
    Qwen3.5-27B & 41.3 & 33.3 & 41.2 & 52.3 & 57.9 & H2$\rightarrow$H3 & 30.4 \\
    Qwen3.5-35B-A3B & 39.3 & 31.1 & 40.7 & 53.1 & 57.4 & H2$\rightarrow$H3 & 26.5 \\
    Qwen3.5-122B-A10B & 33.8 & 33.3 & 42.6 & 54.9 & 60.1 & H2$\rightarrow$H3 & 29.1 \\
    \bottomrule
  \end{tabular}
\end{table}

\paragraph{Gemma 4 replication.}
\label{app:gemma4_ssa}
Gemma 4 is used as a cross-family replication, not as a separate mechanism claim. The intervention profile is S2/S3-centered: E2B/E4B peak at H2$\rightarrow$H3, while 26B-A4B and 31B peak at H1$\rightarrow$H2. This supports the behavioral conclusion that verified intermediate structure helps only when it approaches rule-to-instance alignment.

\begin{figure}[H]
  \centering
  \includegraphics[width=0.82\textwidth]{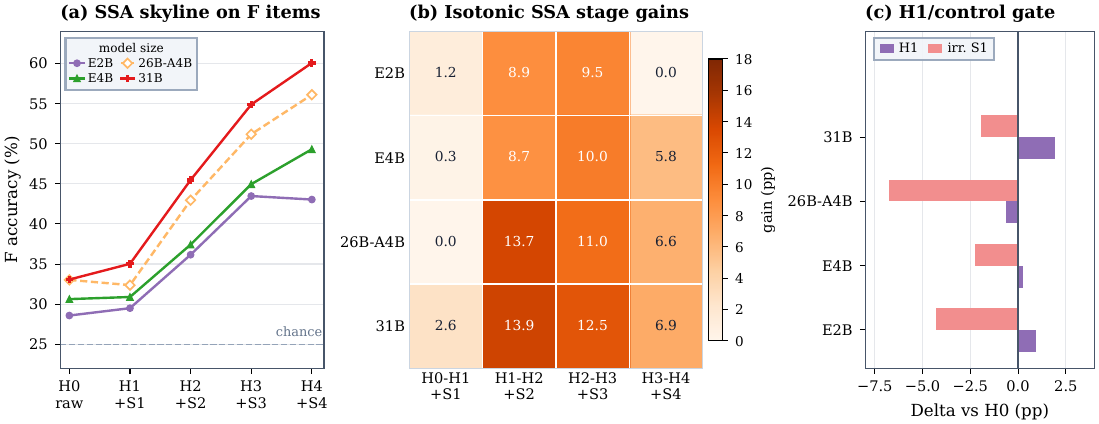}
  \caption{Gemma~4 \ssa{} replication. The cross-family profile is appendix-only and is used to check whether verified intermediate structure helps beyond the Qwen3.5 family.}
  \label{fig:app_stage_ssa_gemma4}
\end{figure}

\subsection{Paired Direct vs.\ Thinking Raw Values and Error Audit}
\label{app:thinkgain_scatter}
The matched thinking comparison uses ten direct/thinking pairs. Thinking improves P on nine of ten rows, while R and F drop on all ten under this protocol. This supports the bounded claim that longer traces are not a reliable repair for rule-to-instance binding; it does not imply that all forms of deliberation are useless.

\definecolor{clPos}{RGB}{187,247,208}
\definecolor{clNeg}{RGB}{254,202,202}
\definecolor{clZero}{RGB}{241,245,249}

\newcommand{\tgpos}[1]{\cellcolor{clPos}$+$#1}
\newcommand{\tgneg}[1]{\cellcolor{clNeg}$-$#1}

\begin{table}[t]
  \centering
  \small
  \setlength{\tabcolsep}{6pt}
  \renewcommand{\arraystretch}{1.12}
  \caption{%
    \textbf{Compact \tg{} deltas for explicit thinking controls.}
    $\Delta$ = thinking$-$non-thinking accuracy (pp).
    \textbf{@P}: perception-probe level;
    \textbf{@R}: rule-probe level;
    \textbf{@F}: full-item level.
    Main leaderboard rows already show the paired raw scores; this
    appendix table reports only the deltas to avoid duplicating the
    leaderboard.%
  }
  \label{tab:thinkgain}
  \begin{tabular}{@{}llccc@{}}
    \toprule
    & \textbf{Pair} & \textbf{@P} & \textbf{@R} & \textbf{@F} \\
    \midrule
    T1 & GPT-5.4 \{direct $\to$ thinking\}
      & \tgpos{2.4} & \tgneg{1.8} & \tgneg{4.6} \\
    T2 & Gemini-3.1-Pro \{direct $\to$ thinking\}
      & \tgpos{1.7} & \tgneg{2.9} & \tgneg{4.8} \\
    T3 & Claude-Opus-4.7 \{direct $\to$ thinking\}
      & \tgpos{2.8} & \tgneg{1.4} & \tgneg{4.0} \\
    T4 & Qwen3.5-Plus \{direct $\to$ thinking\}
      & \tgpos{1.6} & \tgneg{3.8} & \tgneg{6.8} \\
    T5 & grok-4.2-beta \{direct $\to$ thinking\}
      & \tgpos{2.1} & \tgneg{1.1} & \tgneg{3.9} \\
    T6 & Qwen3.5-27B \{direct $\to$ thinking\}
      & \tgpos{1.4} & \tgneg{4.8} & \tgneg{7.2} \\
    T7 & Qwen3.5-9B \{direct $\to$ thinking\}
      & \tgpos{1.1} & \tgneg{5.5} & \tgneg{7.8} \\
    T8 & Qwen3.5-4B \{direct $\to$ thinking\}
      & \tgpos{0.8} & \tgneg{4.9} & \tgneg{8.4} \\
    T9 & Qwen3.5-2B \{direct $\to$ thinking\}
      & \tgpos{0.2} & \tgneg{5.5} & \tgneg{8.7} \\
    T10 & Qwen3.5-0.8B \{direct $\to$ thinking\}
      & \tgneg{0.6} & \tgneg{6.2} & \tgneg{9.1} \\
    \bottomrule
  \end{tabular}
\end{table}

\begin{figure}[H]
  \centering
  \includegraphics[width=0.86\textwidth]{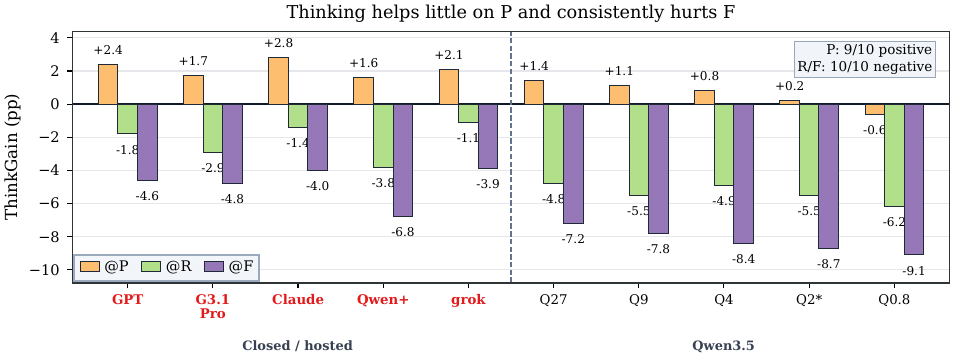}
  \caption{Compact visualization of paired direct-vs-thinking deltas from Table~\ref{tab:thinkgain}. Under this protocol, thinking usually raises P but lowers R and F, so it is not a stable rule-to-instance binding repair.}
  \label{fig:app_thinkgain_from_table}
\end{figure}

\paragraph{Token budget sanity check.} A natural worry about the paired comparison is that thinking traces are clipped before they reach the answer tag, so the R and F drops would reflect truncation rather than a binding effect. We use a fixed decoding budget of 32{,}768 output tokens for every direct and thinking row, and report mean, p95, and cap-hit rate (share of items whose output reaches the cap) per model and mode in Tab.~\ref{tab:token_budget}. p95 stays below the cap on every row, the cap-hit rate never exceeds 5.4\%, and thinking already receives 10--25$\times$ the direct token budget. Verbosity also varies by more than $2\times$ across thinking rows (mean 5.9k--13.2k tokens), yet R and F fall on every paired row, so the drop is not driven by a single overly long or overly short trace. Re-computing the paired \tg{} deltas after excluding cap-hit items leaves the sign and ordering of @P, @R, and @F in Tab.~\ref{tab:thinkgain} unchanged. Truncation and trace-length heterogeneity are therefore not viable explanations for the observed thinking pattern.

\begin{table}[t]
  \centering
  \small
  \setlength{\tabcolsep}{5pt}
  \renewcommand{\arraystretch}{1.06}
  \caption{Output-token statistics for the paired direct/thinking rows in Tab.~\ref{tab:thinkgain}. The decoding cap is 32{,}768 tokens; \textit{cap\%} is the share of items whose output reaches the cap. p95 stays below the cap on every row and cap\% never exceeds 5.4\%, so thinking traces are not truncated. Thinking already receives 10--25$\times$ the direct token budget, yet R and F fall on every paired row, which means the drops in Fig.~\ref{fig:thinking_paired_bars_all} cannot be attributed to budget ceilings.}
  \label{tab:token_budget}
  \begin{tabular}{@{}lrrrrrr@{}}
    \toprule
     & \multicolumn{3}{c}{Direct} & \multicolumn{3}{c}{Thinking} \\
    \cmidrule(lr){2-4} \cmidrule(lr){5-7}
    Model & mean & p95 & cap\% & mean & p95 & cap\% \\
    \midrule
    Qwen3.5-0.8B        & 612 & 2{,}104 & 1.8 & 11{,}244 & 22{,}890 & 4.2 \\
    Qwen3.5-2B          & 583 & 1{,}987 & 1.5 & 12{,}876 & 23{,}104 & 5.1 \\
    Qwen3.5-4B          & 548 & 1{,}842 & 1.2 & 13{,}205 & 23{,}456 & 5.4 \\
    Qwen3.5-9B          & 522 & 1{,}763 & 0.9 & 11{,}892 & 22{,}018 & 3.7 \\
    Qwen3.5-27B         & 498 & 1{,}654 & 0.7 & 10{,}547 & 20{,}893 & 2.4 \\
    \midrule
    Qwen3.5-35B-A3B     & 568 & 1{,}893 & 1.3 & 12{,}105 & 22{,}567 & 4.3 \\
    Qwen3.5-122B-A10B   & 462 & 1{,}547 & 0.4 &  8{,}742 & 18{,}632 & 1.2 \\
    Qwen3.5-Plus        & 424 & 1{,}432 & 0.3 &  8{,}156 & 17{,}893 & 0.9 \\
    \midrule
    Claude-Opus-4.7     & 547 & 3{,}198 & 0.0 &  6{,}824 & 18{,}920 & 0.3 \\
    GPT-5.4             & 439 & 1{,}153 & 0.0 &  5{,}918 & 17{,}204 & 0.2 \\
    grok-4.2-beta       & 352 & 2{,}217 & 0.0 &  7{,}235 & 19{,}108 & 0.4 \\
    gemini-3.1-pro      & 545 & 3{,}186 & 0.0 &  6{,}512 & 18{,}347 & 0.3 \\
    \bottomrule
  \end{tabular}
\end{table}

\paragraph{Error audit of thinking traces.}
\label{app:thinking_audit}
The paired deltas show that thinking lowers R and F, but not which errors it adds. We audit the failures directly on a sample of 180 stems, 20 per operation. For every paired row in Table~\ref{tab:thinkgain} we take the direct trace and the thinking trace of each stem, which gives 1{,}800 F items per condition. Each trace is labeled with its first decisive error, blind to the condition it came from. The error types are: \emph{correspondence drift} (the trace states the correct rule but applies it to the wrong slot or instance); \emph{rule mutation} (the trace states a rule and later rewrites it); \emph{distractor capture} (a near-miss option pulls the answer); \emph{late answer switching} (the trace reaches the correct option and then switches at the last step); \emph{perception error} (an S1 failure); and \emph{rule absent} (no rule is stated at any point). Table~\ref{tab:thinking_audit} reports errors per 100 F items.

\begin{table}[htbp]
  \centering
  \small
  \setlength{\tabcolsep}{6pt}
  \renewcommand{\arraystretch}{1.08}
  \caption{First decisive error per 100 F items on 180 stems $\times$ 10 paired rows (1{,}800 F items per condition), labeled blind to condition.}
  \label{tab:thinking_audit}
  \begin{tabular}{@{}lrrr@{}}
    \toprule
    Error type & Direct & Thinking & $\Delta$ \\
    \midrule
    Correspondence drift & 13.8 & \textbf{22.4} & \textbf{$+$8.6} \\
    Rule mutation & 7.2 & 10.9 & $+$3.7 \\
    Distractor capture & 9.6 & 10.2 & $+$0.6 \\
    Late answer switching & 3.6 & 4.1 & $+$0.5 \\
    Perception error (S1) & 9.0 & 6.8 & $-$2.2 \\
    Rule absent throughout & 13.2 & 10.9 & $-$2.3 \\
    Other & 3.6 & 2.7 & $-$0.9 \\
    \bottomrule
  \end{tabular}
\end{table}

The change is selective. Perception errors and rule-absent traces both fall, while the increase sits on the rule-to-instance side: correspondence drift rises by 8.6 and rule mutation by 3.7 per 100 items. Distractor capture and late answer switching stay nearly flat, so the extra errors are neither near-miss option pull nor last-step answer flipping. Rule mutation is the F-side counterpart of the R drop in Table~\ref{tab:thinkgain}.

\paragraph{Drift and trace length.} Within the thinking condition we split the audited traces into equal thirds by reasoning length. Correspondence drift per 100 F items is 15.1, 21.8, and 30.4 from the shortest to the longest third, against 13.8 for direct traces. The shortest third is close to direct, and the longest third drifts twice as often.

\subsection{Auxiliary Audits and Sanity Checks}
\label{app:aux_metrics}
\label{app:stc_audit}
\label{app:dataset_sanity}
\label{app:contamination}
\label{app:shortcut}
Auxiliary metrics are validity and robustness checks, not separate headline findings. \stc{} tests whether P/R/F non-redundancy adds information beyond final accuracy; \vg{} tests image dependence; lucky-guess identifies cases where F is correct despite earlier stage failure. Dataset-level sanity audits (image integrity, duplicate stems, malformed probes, shortcut pre-checks via text-only / caption-only / parse-only conditions~\citep{agrawal2018,geirhos2020}, and pHash + CLIP deduplication) are passed at the time of release; complete logs accompany the release rather than the appendix.

\begin{table}[htbp]
  \centering
  \small
  \setlength{\tabcolsep}{5pt}
  \renewcommand{\arraystretch}{1.08}
  \caption{Auxiliary audit summary.}
  \label{tab:aux_audit_summary}
  \begin{tabular}{@{}p{0.20\textwidth}p{0.32\textwidth}p{0.22\textwidth}p{0.16\textwidth}@{}}
    \toprule
    Audit & Purpose & Main result & Use in paper \\
    \midrule
    \stc{} & Test P/R/F non-redundancy & 0.0--6.0\% across full rows & Supplementary \\
    \vg{} & Test image dependence & full-image protocol retained & Sanity check \\
    Lucky guess & Detect correct F with broken trace & tracked in L2 logs & Sanity check \\
    Shortcut audit & Text-only / caption-only / parse-only & Pass & Construction \\
    pHash + CLIP dedup & Released stems vs.\ public sources & Pass & Construction \\
    Dual-VLM pre-screen & Frozen VLM solvers as leakage probe & Pass & Construction \\
    \bottomrule
  \end{tabular}
\end{table}

\FloatBarrier

\section{Qualitative Case Study Gallery}
\label{app:case}
This section is the appendix's main bulk. The remaining manual figures focus on case-error evidence. Each error card should include: (i) the operation label, (ii) the stem and relevant probes, (iii) the model name, model-selected answer, ground-truth answer, and a short output excerpt, (iv) the judge's S1--S4 stage marks where applicable, (v) the failure type, and (vi) a one-line note explaining why the case matters.

\subsection*{Case Index}
\phantomsection
\label{app:case_index}

\begin{table}[H]
  \centering
  \scriptsize
  \setlength{\tabcolsep}{3pt}
  \renewcommand{\arraystretch}{1.0}
  \caption{Index of qualitative case figures. Each row corresponds to one case figure rendered manually using the main-text RI/VP and P/R/F palette.}
  \label{tab:case_index}
  \begin{tabular}{@{}llllp{0.30\textwidth}@{}}
    \toprule
    Case ID & Operation & P/R/F outcome & Failure type & What it shows \\
    \midrule
    E.1 & RI-Pos & P\checkmark R\checkmark F\checkmark & solved reference & success baseline for reading the cards \\
    E.2 & VP-View & P$\times$ R$-$ F$\times$ & case error: can't see & P-probe outputs miss option-B visual elements \\
    E.3 & RI-Attr & P\checkmark R$\times$ F$\times$ & case error: can't reason & model outputs infer the wrong grouping rule \\
    E.4 & RI-Qty & P\checkmark R\checkmark F$\times$ & case error: can't bind & correct count rule, wrong answer slot \\
    E.5 & VP-View & worked S1--S4 trace & hard solved trace & multi-view projection and option elimination \\
    E.6 & RI-Sty & direct \checkmark{} / thinking $\times$ & direct vs.\ thinking & longer reasoning drifts to the wrong rule \\
    \bottomrule
  \end{tabular}
\end{table}

\subsection{Solved Reference Case (E.1)}
\label{app:case_solved}
\begin{figure}[H]
  \centering
  \includegraphics[width=0.96\textwidth,height=0.80\textheight,keepaspectratio]{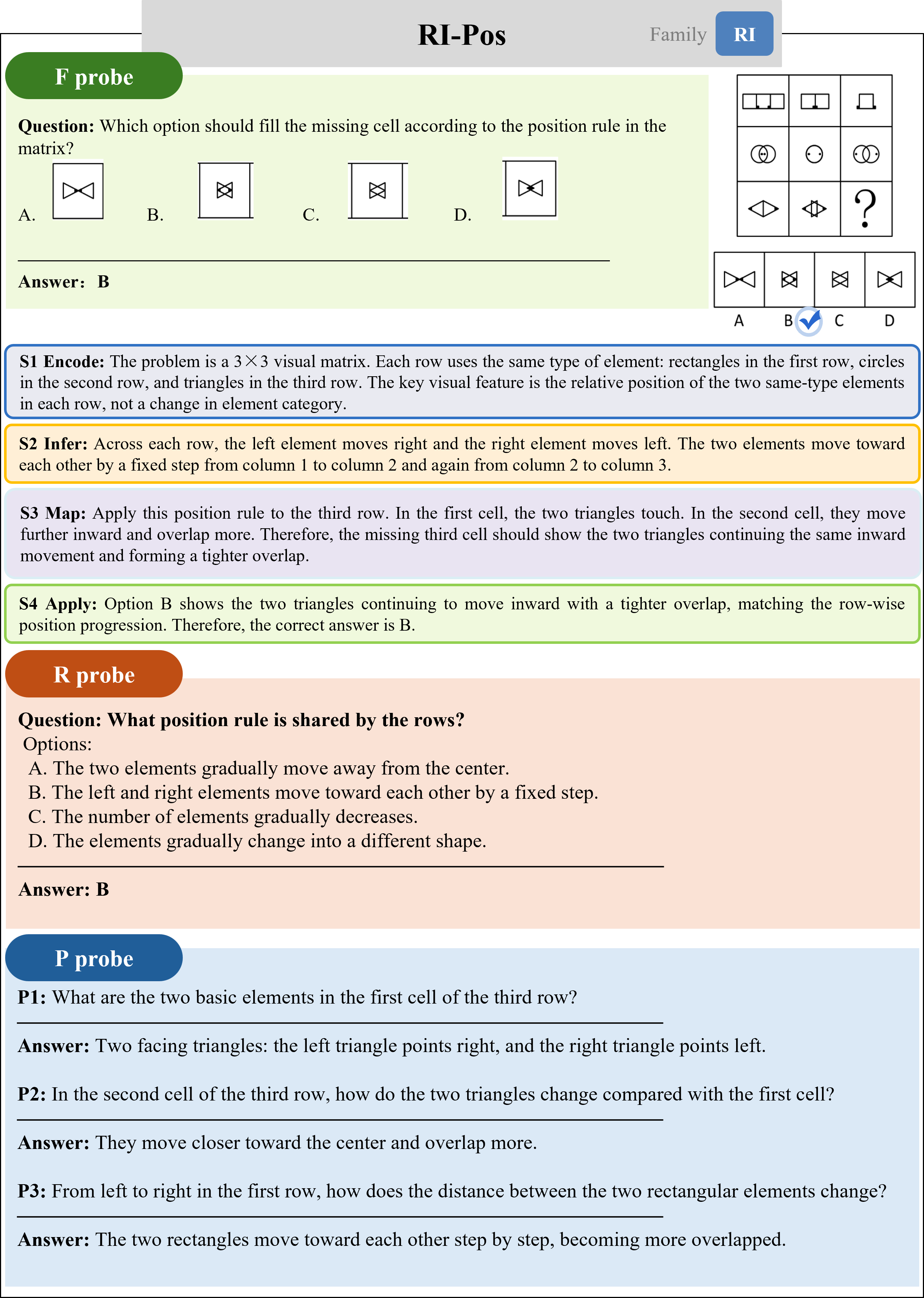}
  \caption{Solved RI-Pos reference case. The figure shows a fully resolved position-reasoning example with the F, R, and P probes, the selected correct option, and the corresponding S1--S4 process annotation.}
  \label{fig:case_E1_solved}
\end{figure}

\subsection{Case-Error: Can't-See (E.2)}
\label{app:case_cant_see}
\begin{figure}[H]
  \centering
  \includegraphics[width=0.96\textwidth,height=0.80\textheight,keepaspectratio]{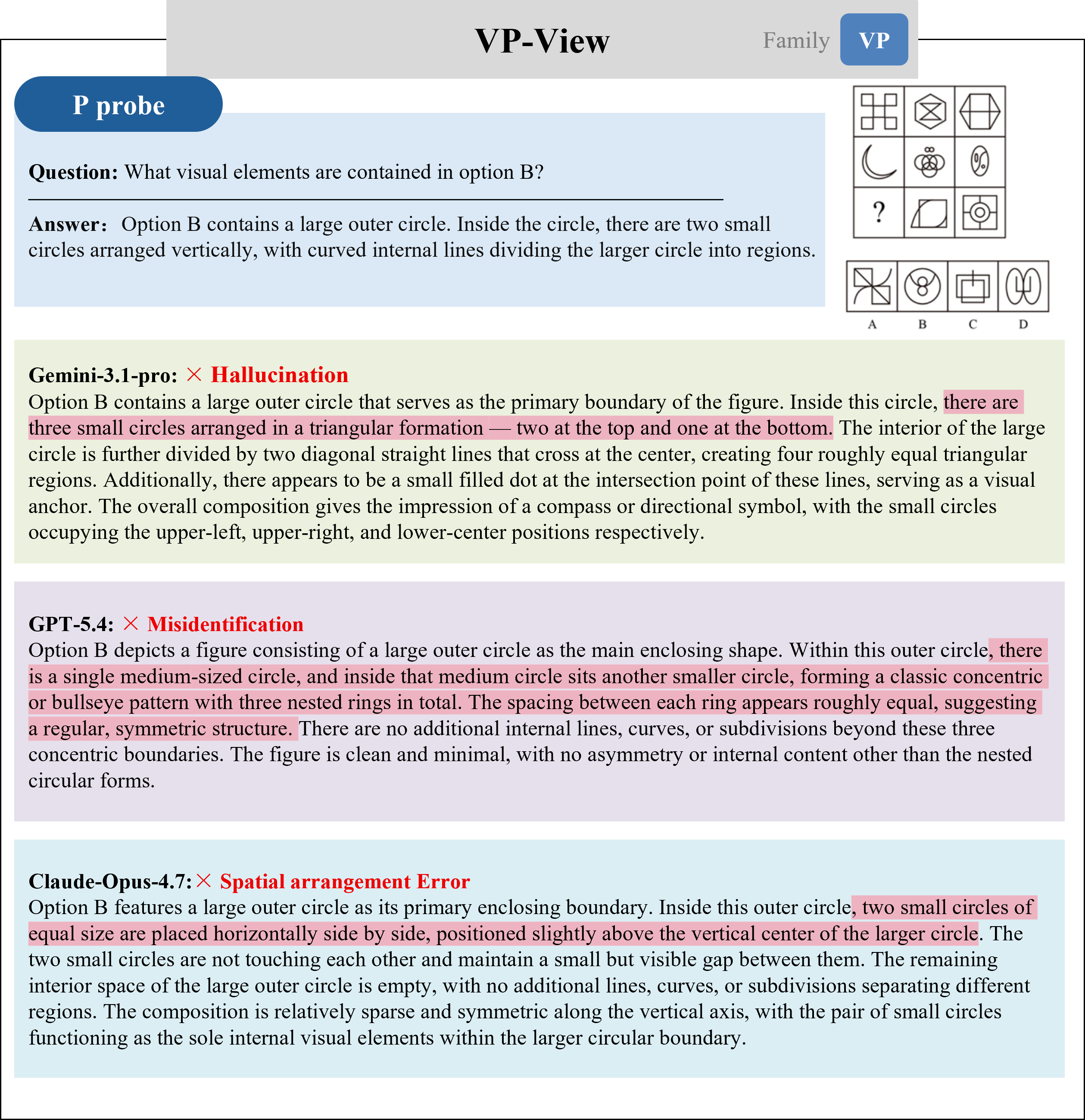}
  \caption{Can't-see P-probe case. The figure illustrates VP-View perception errors on option B: the model outputs hallucinate extra circles, misidentify the nested structure, or misplace the two inner circles. The error is localized to S1 visual encoding rather than rule induction or rule-to-instance binding.}
  \label{fig:case_E2_cant_see}
\end{figure}

\subsection{Case-Error: Can't-Reason (E.3)}
\label{app:case_cant_reason}
\begin{figure}[H]
  \centering
  \includegraphics[width=0.96\textwidth,height=0.80\textheight,keepaspectratio]{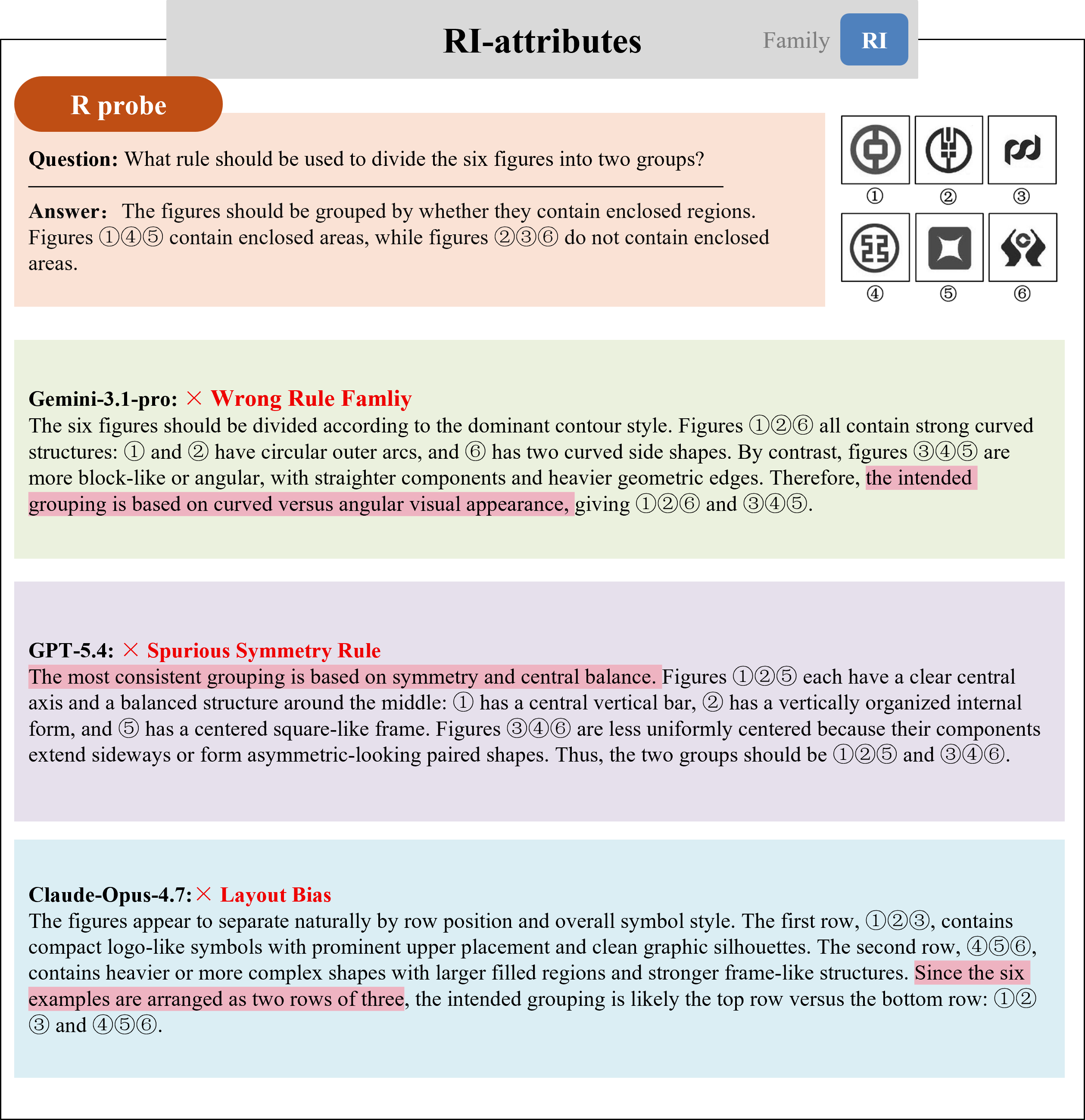}
  \caption{Can't-reason R-probe case. The figure illustrates RI-Attr rule-induction errors: the models refer to visible symbol properties, but replace the ground-truth enclosed-region rule with spurious grouping rules such as contour style, symmetry, or display layout. The error is localized to S2 rule inference rather than S1 visual encoding or S3 rule-to-instance binding.}
  \label{fig:case_E3_cant_reason}
\end{figure}

\subsection{Case-Error: Can't-Bind (E.4, main appendix evidence)}
\label{app:case_cant_bind}
This sub-section is the qualitative core of the appendix because it directly visualizes the strict Binding Gap reported in Sec.~\ref{app:binding_gap_ci}. Each card shows that the model has both the right perception and the right rule on the same stem, yet binds the rule to the wrong answer slot.

\begin{figure}[H]
  \centering
  \includegraphics[width=0.96\textwidth,height=0.80\textheight,keepaspectratio]{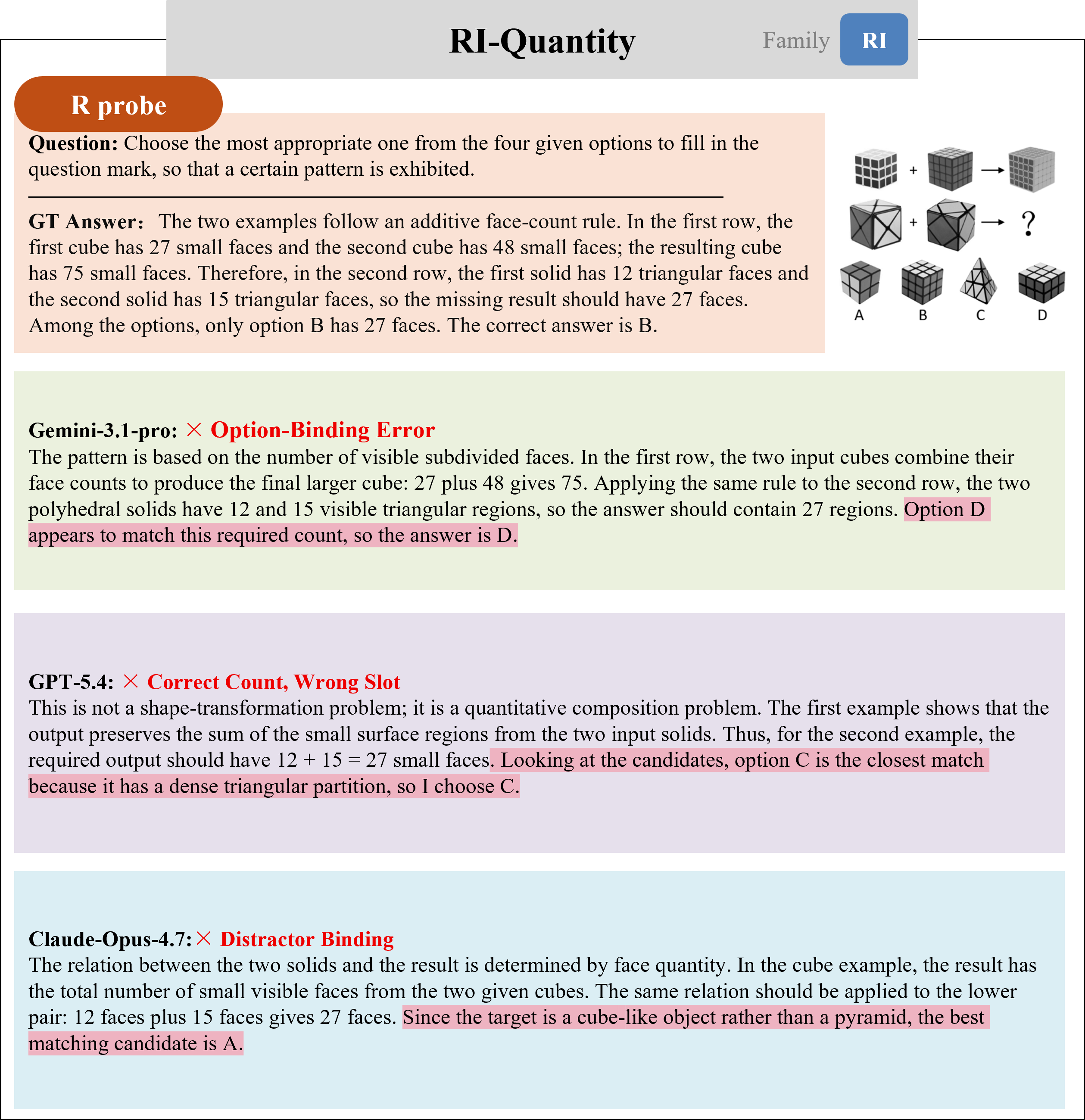}
  \caption{Can't-bind RI-Quantity case. The figure illustrates a quantity-binding failure: the models identify the additive face-count rule and the required target count of 27 faces, but then bind that count to the wrong answer option. The error is localized to S3 rule-to-option mapping rather than S1 visual encoding or S2 rule induction.}
  \label{fig:case_E4_cant_bind}
\end{figure}

\subsection{Hard S1--S4 Worked Trace (E.5)}
\label{app:case_worked_trace}
\label{app:worked_example}
Worked traces show how a full solution is decomposed into S1--S4 stage targets. The VP-View case below is a hard three-view reasoning example: it requires encoding the two adjacent 3D solids, inferring view-consistency constraints, mapping the projected points across candidate top and left views, and applying those constraints to eliminate the distractors.

\begin{figure}[H]
  \centering
  \includegraphics[width=0.96\textwidth,height=0.80\textheight,keepaspectratio]{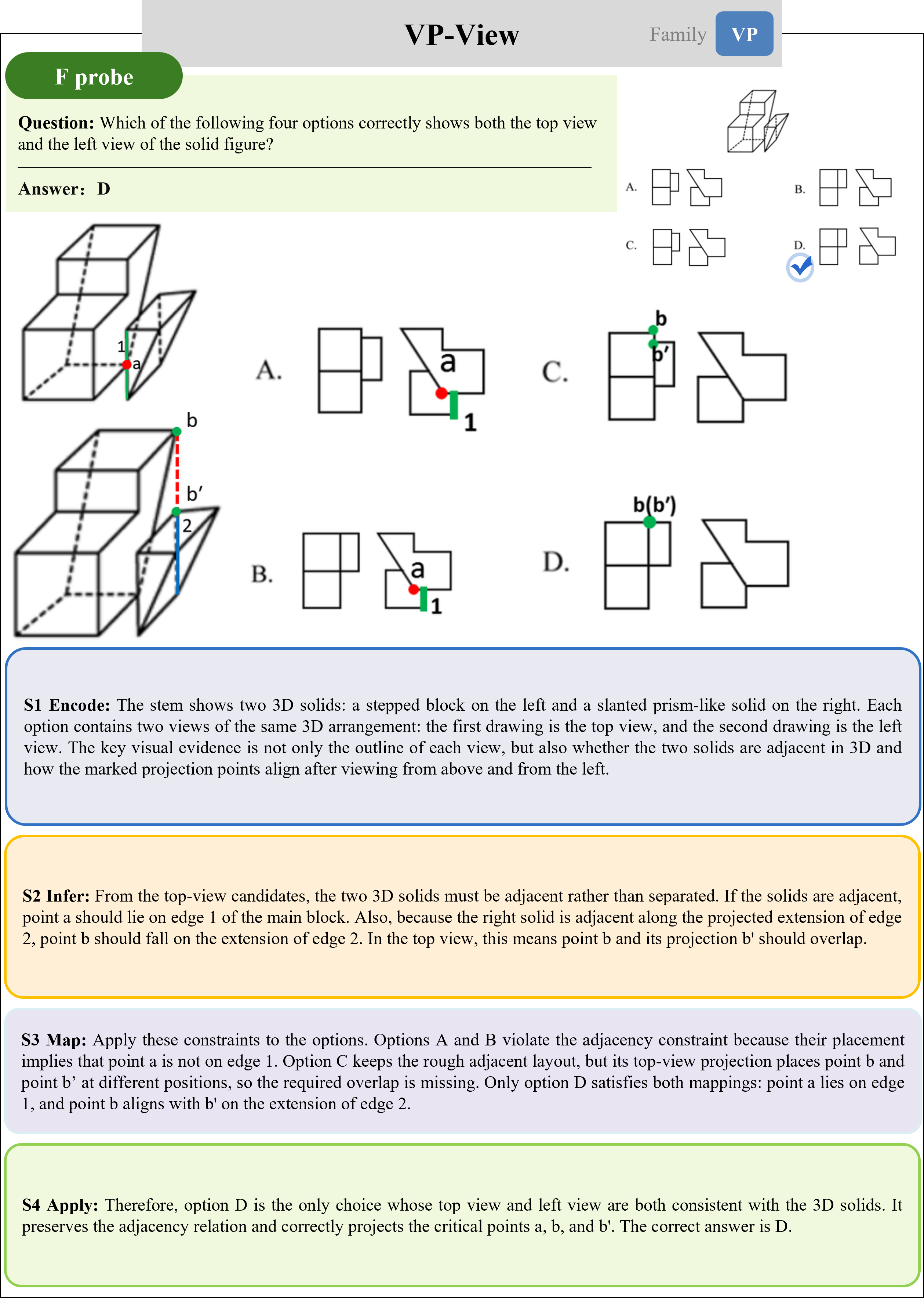}
  \caption{Hard VP-View worked trace. The case illustrates how a three-view matching problem decomposes into S1 visual encoding, S2 view-consistency rule inference, S3 projection mapping, and S4 option elimination.}
  \label{fig:case_E5_worked_trace}
\end{figure}

\subsection{Case-Error: Direct vs.\ Thinking Pairs (E.6)}
\label{app:case_thinking}
Thinking cases visualize the paired \tg{} effect from a single stem perspective: longer reasoning can change the stated rule and does not provide a stable repair for the final answer. Appendix~\ref{app:thinking_audit} measures how often such errors occur on a 1{,}800-item audit. The case below shows a direct response that correctly applies the row-wise overlay rule, while the thinking response drifts to a local triangle-position transformation and selects the wrong option.

\begin{figure}[H]
  \centering
  \includegraphics[width=0.96\textwidth,height=0.80\textheight,keepaspectratio]{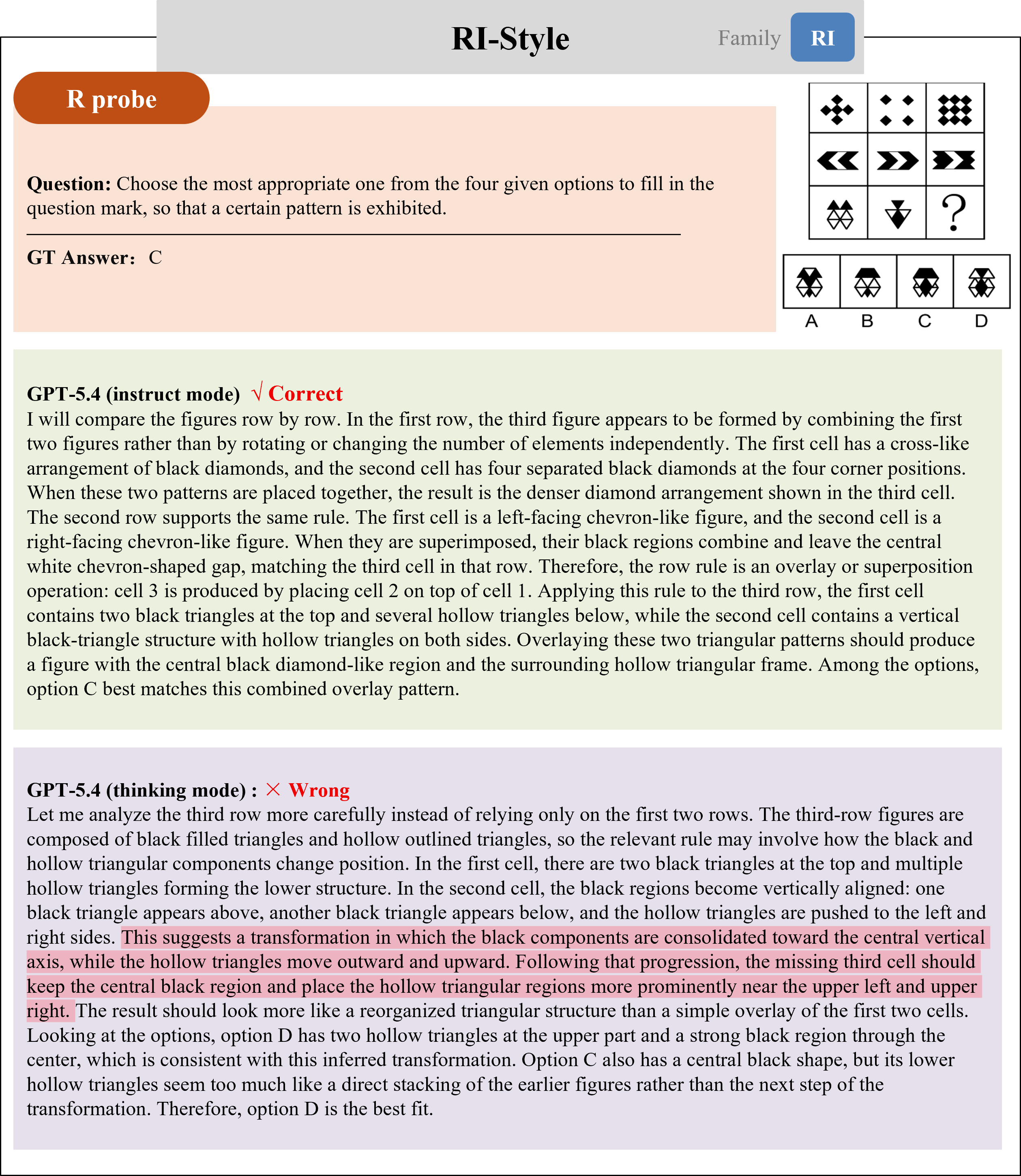}
  \caption{Illustrative direct-vs-thinking RI-Style case. The direct response correctly applies the row-wise overlay rule and selects option C. The thinking response over-analyzes local black and hollow triangle positions, drifts to an incorrect transformation rule, and selects option D. This paired case shows that longer reasoning can destabilize the correct rule rather than repair the final answer.}
  \label{fig:case_E6_thinking}
\end{figure}

\FloatBarrier
\section{Extended Related Work}
\label{app:related_extended}

This appendix expands the literature lists summarized but not enumerated in \S\ref{sec:related}.

\paragraph{Additional MLLM and AVR benchmarks.} Beyond the representative works in the main text, broader MLLM evaluation also includes ScienceQA~\citep{scienceqa}, M3CoT~\citep{m3cot}, MMECoT~\citep{mmecot}, and ENCBench~\citep{encbench}, which mix visual reasoning with science or domain knowledge. Additional knowledge-light AVR benchmarks include VisuRiddles~\citep{visuriddles}, VRIQ~\citep{vriq}, IQBench~\citep{iqbench}, and MORSE-500~\citep{morse500}. They share the AVR motivation but evaluate at the final-answer level only, without shared-stem perception, rule, and full-item probes.

\paragraph{Multi-image and multi-context benchmarks.} A growing line of benchmarks~\citep{mmrb,omibench,mmiu,mibench,mirb,blink,remi,mantis,mementos,mcbench,muirbench} stresses cross-image grounding, multi-context aggregation, or temporal reasoning rather than AVR-specific shared-stem diagnosis. They are complementary to \avrb{} but do not isolate rule-to-instance binding on a single visual stem.

\paragraph{Vision-aware reasoning methods.} Methods that interleave perception with reasoning, such as Thinking with Images~\citep{thinkingimages}, slow perception~\citep{slowperception}, and CogFlow~\citep{cogflow}, focus on image-based deliberation, stepwise perception, or perception--reasoning flows on natural images, videos, or math problems. \avrb{} provides a complementary diagnostic testbed where perception, rule, and full-item probes share the same visual stem and where stage-level interventions (\ssa{}) can be applied.

\section{Limitations and Broader Impact}
\label{app:limitations_impact}

\subsection{Limitations}
\label{app:limitations}
\label{app:appendix_limitations}
\avrb{} is depth-first rather than breadth-first: 2{,}298 stems and 19{,}533 tasks support the shared-stem P/R/F design and S1--S4 annotation, but at a smaller absolute scale than purely automatic AVR datasets. The benchmark is currently English-only and uses a finite operation taxonomy, so performance should not be overgeneralized to visual reasoning as a whole. The full L2 and \ssa{} diagnostics primarily cover the Qwen3.5 family, with Gemma 4 used as a cross-family replication rather than an independent mechanism claim. All findings are behavioral diagnostics: they characterize where models break down on shared-stem evidence, but they do not identify an internal binding module. L3 AttrTag is retained as metadata and is not treated as evidence for the current claims. Per-item L2 stage labels depend on judge calibration. S3 is also the hardest stage for human annotators to agree on, so per-item S3 labels should be read as noisy, while the aggregate S3-centered pattern is supported by the joint L2, human re-annotation, and \ssa{} evidence.

\subsection{Broader Impact}
\label{app:broader_impact}
\avrb{} pushes AVR evaluation away from a single ``answer-correct'' leaderboard and toward locating where solution paths break. The shared-stem and process-level protocol can be misused to over-state ``general reasoning ability''; we therefore emphasize that \avrb{} measures rule-to-instance binding on weak-knowledge visual stems and is not a test of general intelligence. Released images are abstract, provenance-screened, and contain no personal or sensitive content; the hidden test set additionally uses option shuffling and answer-mapping randomization to reduce contamination risk. Users adopting \avrb{} for model selection should report the full P/R/F profile, not only F accuracy, so that perception, rule, and binding axes are visible to downstream stakeholders.

\end{document}